\documentclass[11pt]{article}
\usepackage[final]{acl}

\usepackage{times}
\usepackage{latexsym}
\usepackage{xcolor}
\usepackage[most]{tcolorbox}
\usepackage{pifont}
\usepackage{makecell}
\usepackage{amsmath,amsfonts,amssymb}
\usepackage{textcomp} % Required for some IPA symbols if using tipa
\usepackage{tipa}     % Required for IPA phonetic symbols
\usepackage{cleveref}
\usepackage{caption}
\crefname{appendix}{Appendix}{Appendices}
\Crefname{appendix}{Appendix}{Appendices}

\usepackage[T1]{fontenc}
\usepackage{tabularx}
\usepackage{array}
\usepackage{multirow}
\usepackage{booktabs}
\usepackage[utf8]{inputenc}
\DeclareUnicodeCharacter{1EA1}{\d a}

\usepackage{microtype}

\usepackage{inconsolata}

\usepackage{graphicx}
\usepackage{lipsum}
\usepackage{tikz}
\usetikzlibrary{arrows.meta,positioning}

\usepackage{fontawesome5}
\usepackage{paralist}
\usepackage{xspace} % whether to add space

\definecolor{mygreen}{rgb}{0.1, 0.6, 0.1} % A slightly darker, readable green
\newcommand{\cmark}{\textcolor{mygreen}{\ding{51}}}
\newcommand{\xmark}{\textcolor{red}{\ding{55}}}

\newcommand{\taubench}{$\tau^2$-Bench\xspace}
\newcommand{\seatau}{SEATauBench\xspace}

\newcommand{\sctools}{L2 Tools\xspace}

\newcommand{\scdomain}{L2 Domain\xspace}

\title{%
SEATauBench: Progressively Adapting Tool-Agent-User Evaluation \\
Into Low-Resource Southeast Asian Languages
}

\author{
 \textbf{My Chiffon Nguyen\textsuperscript{1}},
 \textbf{Aulia Adila\textsuperscript{1,2}},
 \textbf{Saksorn Ruangtanusak\textsuperscript{1,3}},
\\
 \textbf{Kittiphat Leesombatwathana\textsuperscript{1,4,5}\thanks{
  Most of this work was conducted while Kittiphat Leesombatwathana was at Chulalongkorn University.
}},
 \textbf{Vissuta Gunawan Lim\textsuperscript{1,6}},
\\
 \textbf{Patomporn Payoungkhamdee\textsuperscript{1,5}},
 \textbf{Samuel Cahyawijaya\textsuperscript{1,7}}
\\
 \textsuperscript{1}SEACrowd,
 \textsuperscript{2}AI Singapore, Nanyang Technological University,
 \textsuperscript{3}SCB DataX, SCBX Group,
 \\
 \textsuperscript{4}Chulalongkorn University,
 \textsuperscript{5}VISTEC,
 \textsuperscript{6}University of Indonesia,
 \textsuperscript{7}Cohere
 \\
 \texttt{chiffonng136@gmail.com, aulia.adila@aisingapore.org}, \\
 \texttt{saksorn.ruangtanusak@data-x.ai, vissuta.gunawan@ui.ac.id}\\
 \texttt{\{kittphat.l\_s26, patomporn.p\_s21\}@vistec.ac.th}, \texttt{samuelcahyawijaya@cohere.com}
}

\begin{document}
\maketitle
\begin{abstract}

While AI development and evaluation for Southeast Asia (SEA) has grown rapidly, agent capabilities in regional languages are still poorly understood despite its importance to sovereign AI.
To fill this gap, we introduce SEATauBench, the first agent-focused evaluation framework for SEA sovereign AI. It adapts $\tau^2$-Bench~\cite{barres2025tau2} to five languages---Mandarin, Vietnamese, Thai, Indonesian, and Filipino---and evaluates agents across progressively localized settings that vary the language of user-agent interaction, tool specifications, and task domains. Across three models, we find that English agent capabilities transfer reasonably well when only the conversation language changes, but quality and robustness degrade sharply as more task contexts are localized, with the largest losses in full domain adaptation. We also highlight the limits of English-only agent assessment for predicting agent capabilities in SEA languages. More broadly, SEATauBench provides a diagnostic benchmark and reusable adaptation pipeline for building reliable multilingual agents for linguistically diverse regions. Data and code can be accessed at \url{https://github.com/SEACrowd/SEATauBench}\footnote{SEATauBench is pronounced "si-tau-bench", similar to the Filipino word for string beans, "sitaw".}.

\end{abstract}

\section{Introduction}

\begin{figure}[!t]
    \centering
    \includegraphics[width=\linewidth]{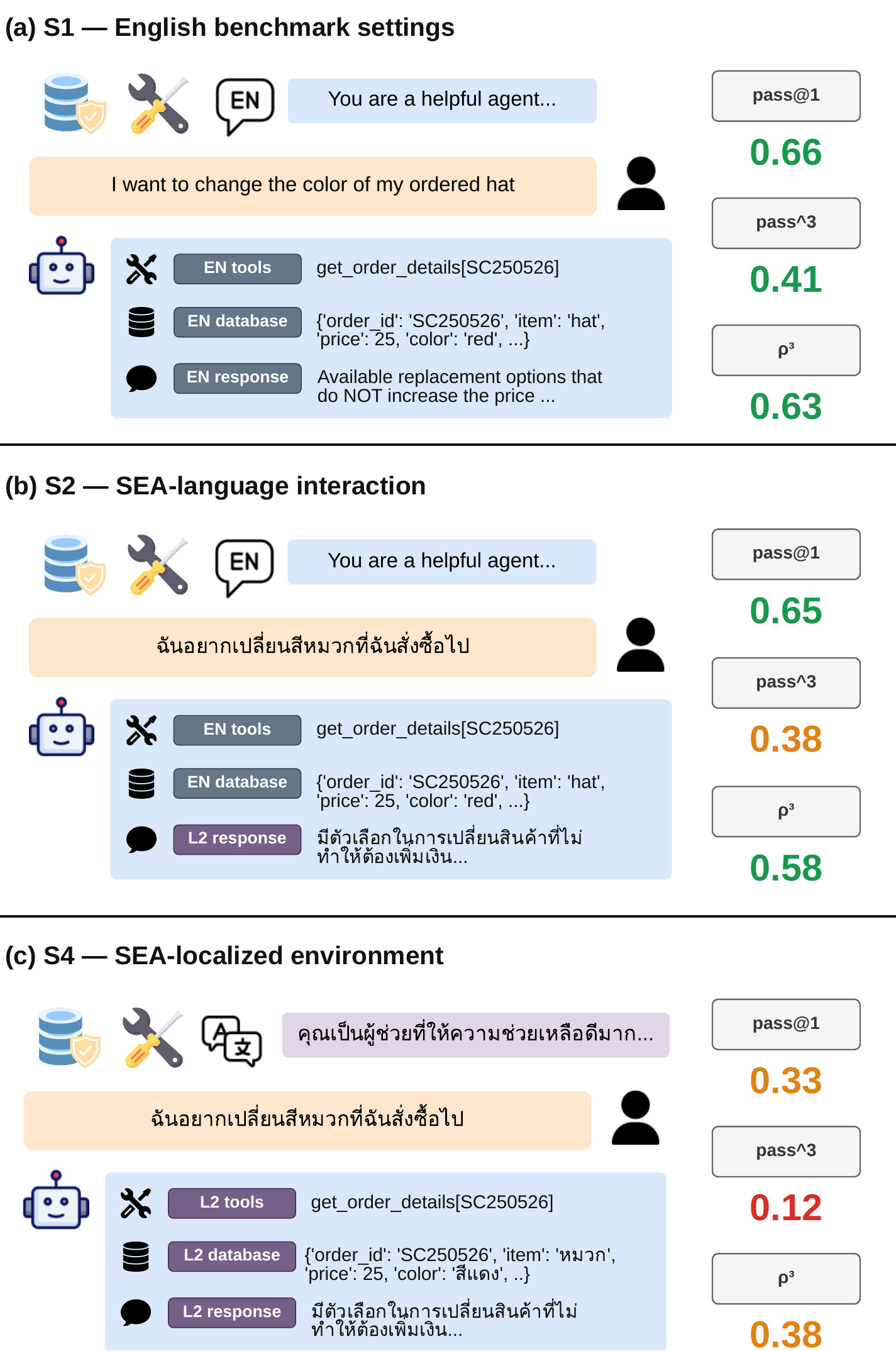}
    \caption{Illustration of \seatau's progressive localization design. The same task is evaluated while dialogue, tool interfaces, and domain context are localized one surface at a time.}
    \vspace{-5mm}
    \label{fig:progressive-localization}
\end{figure}

\begin{table*}[!th]
\centering
\resizebox{\linewidth}{!}{%
\begin{tabular}{@{} l l l c l l c @{}}
    \toprule
    \textbf{Benchmark} & \makecell{\textbf{Language}\\\textbf{coverage}} & \textbf{Interaction} & \makecell{\textbf{Simulated}\\\textbf{user}} & \makecell{\textbf{Action/state}\\\textbf{grounding}} & \makecell{\textbf{Localization}\\\textbf{depth}} & \makecell{\textbf{Multiple}\\\textbf{runs}} \\
    \midrule
    MASSIVE-Agents~\citep{kulkarni2025massiveagents} & 52 & Single-turn & \xmark & Function calls; AST & User utterances; English names & \xmark \\
    Ticket-Bench~\citep{almeida2025ticketbenchkickoffmultilingualregionalized} & 6 & Single-query & \xmark & Ticket functions; final state & Queries, entities, functions & \cmark \\
    X-WebAgentBench~\citep{wang-etal-2025-x} & 14 & Interactive web steps & \xmark & Search/click; product catalog & Instructions, products, web UI & \xmark \\
    WebMMU~\citep{awal-etal-2025-webmmu} & 4 & Offline task instances & \xmark & UI grounding; code edits & Web content, prompts & \xmark \\
    MAPS~\citep{hofman-etal-2026-maps} & 12 & Mixed agent loops & \xmark & Native tools; task executors & Instructions only & \cmark \\
    macOSWorld~\citep{macosworld} & 5 & Interactive GUI steps & \xmark & Mouse/keyboard; OS state & Instructions, OS UI & \xmark \\
    AgentClinic~\citep{schmidgall2025agentclinicmultimodalagentbenchmark} & 7 & Multi-turn & \cmark & Exam/imaging tools; diagnosis & Patient cases, dialogue & \xmark \\
    \textbf{\seatau{} (Ours)} & 6 & Multi-turn & \cmark & API tools; final DB & Dialogue, tools, policies, DB & \cmark \\
    \bottomrule
    \end{tabular}%
}
\caption{Multilingual agent benchmarks compared with \seatau. All benchmarks with multiple run evidence report three trials per task. \seatau is the only benchmark here that progressively localizes user dialogue, tool interfaces, and domain context in multi-turn tasks. }
\label{tab:benchmark-comparison}
\end{table*}

Sovereign AI seeks autonomy over digital futures while keeping systems relevant to the languages, institutions, and cultural settings of the people they serve \cite{chae-etal-2025-assessing-sovereign,mushkani2025position,putra2024governingai}. Southeast Asia (SEA) is home to more than 700 million people, yet English-centric development and evaluation provide an incomplete picture of agent capability across the region \cite{ey2026sovereignai,lovenia2024seacrowd}.

Existing benchmarks often isolate these capabilities. A tool-using agent must understand a user's language, follow a domain policy, ground decisions in structured state, invoke tools, and complete tasks over multiple turns. Localizing every surface at once, however, obscures the source of failure.

We introduce \seatau, a multilingual extension of \taubench{}~\cite{barres2025tau2} for five SEA languages: Vietnamese, Indonesian, Thai, Filipino, and Mandarin Chinese. Four controlled scenarios progressively localize the dialogue, tool schemas, and domain context: S1 is the English baseline, S2 localizes the interaction, S3 localizes the tool schemas, and S4 localizes the full task domain. The progression separates language generation from tool grounding and domain reasoning while keeping the underlying tasks and scoring interface aligned.

The benchmark contributes three components: (1) a \textbf{progressive localization benchmark} with four scenarios that isolate dialogue, tool schemas, and domain context; (2) a \textbf{pipeline that separates localized agent-facing context from canonical execution} with automated and human audits; and (3) a \textbf{reliability analysis} showing how quality and robustness across repeated runs vary by language, domain, and localization depth.

Across three agent models, English task success transfers reasonably well when only dialogue is localized, but quality and robustness decline as tools and domain context are localized, with the largest losses in L2 Domain (\Cref{sec:main-pref-degrade}). The decline varies by language, model, and domain, and English performance does not reliably predict repeated-run robustness in SEA languages (\Cref{sec:breakdown,sec:en-proxy}). \seatau therefore serves as a diagnostic benchmark for multilingual, tool-using agents, especially for the SEA region.

\section{Related Work}

\subsection{Agent evaluation}

Task-oriented dialogue benchmarks such as MultiWOZ \cite{budzianowski2018multiwoz} and MASSIVE \cite{fitzgerald2023massive} establish evaluation for goal-directed dialogue and multilingual intent and slot understanding. ToolLLM \cite{qin2024toolllm} and the Berkeley Function Calling Leaderboard \cite{patil2025bfcl} measure tool selection and argument generation, while ToolSandbox \cite{lu2024toolsandbox} adds conversational context, mutable state, and action dependencies.

These benchmarks isolate useful capabilities, but they do not fully test an agent that infers a user's goal, updates external state, and repeats a specialized task across multiple turns. $\tau$-Bench~\cite{yao2024tau} evaluates this setting with a simulated user, domain-specific tools, and final-state metrics. $\tau^2$-Bench~\cite{barres2025tau2} adds dual control for telecom domain. These English-centered benchmarks provide the interaction structure that \seatau extends with controlled multilingual conditions and progressive localization.

\subsection{Multilingual evaluation}

Multilingual resources broaden evaluation beyond English. mMMLU~\cite{hendrycks2021mmmlu} and GlotEval~\cite{luo2025gloteval} measure multilingual knowledge and language understanding. For Southeast Asian languages, NusaX~\cite{winata2023nusax} and NusaWrites~\cite{cahyawijaya2023nusawrites} provide low-resource language resources, while NusaCrowd~\cite{cahyawijaya2023nusacrowd} and SEACrowd~\cite{lovenia2024seacrowd} provide curated multilingual data and benchmark infrastructure. SeaExam and SeaBench~\cite{liu2025seaexam} cover local questions and open-ended tasks; SEA-HELM~\cite{susanto2025seahelm} covers holistic linguistic, cultural, and safety evaluation; and SEA-VL~\cite{cahyawijaya2025crowdsource} and SEA-VQA~\cite{urailertprasert2024seavqa} extend coverage to multimodal data. These resources are essential, but they largely assess static or closed-ended behavior rather than tool-mediated task completion with a simulated user and structured state.

\subsection{Multilingual agent evaluation}

Recent work brings multilingual evaluation closer to interactive agents. \Cref{tab:benchmark-comparison} summarizes the design choices that matter for this setting. MASSIVE-Agents~\cite{kulkarni2025massiveagents} scales function calling to 52 languages, but evaluates single-turn selection and argument prediction. Ticket-Bench~\cite{almeida2025ticketbenchkickoffmultilingualregionalized} localizes entities and functions in a state-based ticket service, but remains single-query. X-WebAgentBench~\cite{wang-etal-2025-x} evaluates multilingual planning in an interactive web environment, and WebMMU~\cite{awal-etal-2025-webmmu} evaluates multilingual web question answering, code editing, and mockup-to-code generation. MAPS~\cite{hofman-etal-2026-maps} translates instructions for established agent benchmarks and evaluates performance and security. macOSWorld~\cite{macosworld} evaluates multilingual GUI agents on macOS, while AgentClinic~\cite{schmidgall2025agentclinicmultimodalagentbenchmark} combines multilingual clinical cases with simulated doctor-patient interaction and tools. These benchmarks are complementary, but each focuses on a particular interaction environment or localized surface.

\seatau fills this gap by placing a simulated user, tools, policies, and structured state in one multilingual loop. It localizes one agent-visible surface at a time while retaining a canonical execution layer, separating conversational transfer, tool grounding, and domain adaptation. Within this comparison, \seatau is, to our knowledge, the first multilingual benchmark to preserve the task-oriented methodology of $\tau
^2$-Bench while controlling localization across dialogue, tools, and domain context. This design also supports repeated-run reliability analysis, which reveals failures that a single successful trial can hide.

\section{Constructing \texorpdfstring{\seatau}{SEATau}}

\taubench{} is an English benchmark for tool, agent, and user interaction in three service domains: retail, airline, and telecom \cite{yao2024tau,barres2025tau2}. Each task exposes the agent to three types of language: dialogue with a simulated user, tool schemas, and domain context such as task instructions, policies, workflows, and database content.

\seatau adapts this setting to four controlled scenarios (\Cref{sec:data-scenarios}) and five target languages (L2): Vietnamese (vi), Indonesian (id), Thai (th), Filipino (tl), and Mandarin Chinese (zh) (\Cref{sec:data-translation-pipeline}).\footnote{Due to limited native speaker review capacity, we do not include other SEA languages, such as Malay and Lao.} L2 henceforth denotes a non-English target language. Rather than translating the benchmark as a single block, we localize each surface visible to the agent separately. This design isolates the effects of multilingual dialogue, tool grounding, and domain reasoning while preserving a common task definition and scoring procedure.

\subsection{Progressive localization scenarios}
\label{sec:data-scenarios}

\begin{table}[!t]
\centering
\resizebox{\linewidth}{!}{
    \begin{tabular}{lccc}
    \toprule
    \textbf{Scenario} /\ \textbf{L2} & \makecell{\textbf{Tools}} & \makecell{\textbf{Dialogue}} & \makecell{\textbf{DB+Policy}} \\
    \midrule
    (S1) English Baseline & \xmark & \xmark & \xmark \\
    (S2) L2 Interaction & \xmark & \cmark & \xmark \\
    (S3) L2 Tool & \cmark & \xmark & \xmark \\
    (S4) L2 Domain & \cmark & \cmark & \cmark \\
    \bottomrule
    \end{tabular}
}
\caption{\seatau's four scenarios. Checkmarks show which agent-visible surfaces are presented in L2; S2 and S3 isolate dialogue and tool schemas, while S4 localizes the task environment.}
\label{tab:scenarios}
\end{table}

\paragraph{S1. English Baseline.} S1 is the English baseline. Dialogue, tool schemas, task instructions, policies, and database content remain in English.

\paragraph{S2. L2 Interaction.} The simulated user and agent communicate in L2, while tools, policies, databases, and task context remain in English. This scenario isolates multilingual dialogue from localized tool use and domain reasoning. Changes relative to S1 can reflect both agent behavior and changes in the simulated user's behavior.

\paragraph{S3. L2 Tool.} The tool schemas shown to the agent are localized into L2. The conversation and domain context remain in English, and tool names and executable implementations remain unchanged. This scenario tests whether the agent can ground tool semantics, parameter descriptions, and allowed values in another language. We test a variant with one L2 for every schema and a mixed variant in which schemas use multiple L2 languages within the same environment during each run.

\paragraph{S4. L2 Domain.} The simulated user and agent communicate in L2, and the tool descriptions, policies, task descriptions, and database content shown to the agent are also in L2. The underlying task definitions, tool behavior, canonical values, and expected final states remain equivalent to the English benchmark. This preserves executable behavior and makes scores comparable across scenarios.

Together, the scenarios measure how each additional localized surface affects task completion. The progression also separates a dialogue language effect from failures that require the agent to interpret localized tools or domain state.

\subsection{L2 adaptation pipeline}
\label{sec:data-translation-pipeline}

\begin{figure*}[!t]
\centering
\includegraphics[width=0.99\linewidth]{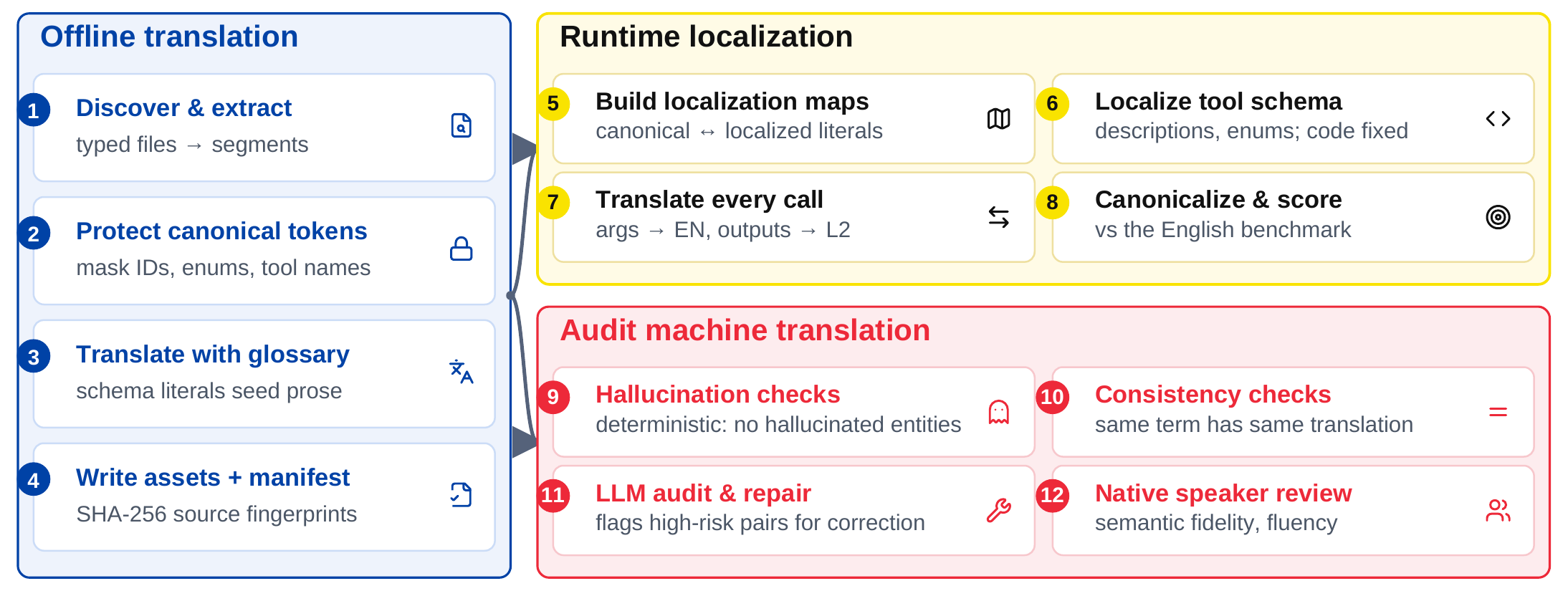}
\caption{The localization pipeline for (S2) \sctools and (S4) \scdomain, as defined in \Cref{tab:scenarios}. Offline translation creates static localized assets, runtime localization preserves canonical execution, and automated and native speaker audits characterize translation quality. We use Gemini 3.1 Flash Lite for offline translation into five target languages.}
\label{fig:pipeline-summary}
\end{figure*}

Comparable scores require localized surfaces to preserve task meaning and executable behavior. We separate the adaptation into an offline stage that creates translated assets and a runtime stage that connects those assets to the canonical benchmark implementation.

\paragraph{Offline translation.} The offline stage uses Gemini 3.1 Flash Lite\footnote{We invoke this model through the LiteLLM route \texttt{vertex\_ai/gemini-3.1-flash-lite}.} to build language-specific assets for the three \taubench{} domains. It extracts natural language spans from tasks, policies, databases, tool docstrings, and schemas, while masking executable tokens such as IDs, status values, tool names, and structural markers. The pipeline translates schema literals first to build a glossary, then translates the remaining spans while preserving original format. It also records a per-language manifest with model metadata and source file SHA-256 fingerprints. Appendix~\ref{app:translated-artifact-stats} reports the resulting artifact statistics.

\paragraph{Runtime localization.} Runtime localization handles content that appears during inference. The wrapper localizes tool schema descriptions, enum choices, default values, and examples in L2 while leaving the implementation unchanged. Before each tool call, it maps localized arguments to canonical English values. After the call, it localizes tool responses and converts final payloads to their canonical forms before scoring. The agent therefore interacts in L2 while the evaluator compares equivalent canonical outcomes across languages.

\subsection{L2 translation audit}
\label{sec:l2-translation-audit}

Translation errors can make a task harder or impossible, alter values needed for execution, or reduce target-language fidelity. We audit these risks in the frozen \emph{experiment snapshot} through solvability analysis, preservation and consistency checks, and native speaker review; approved corrections enter a separate \emph{reviewed release}. 

We first measure empirical achievability. A task is solvable when any of nine runs---three agent models with three trials each---is successful with reward 1. \Cref{tab:solvability-results} shows near-complete solvability in Telecom but larger cross-language gaps in Airline and Retail. Inspection of the 16 tasks solved in English but not in L2 points mainly to incomplete multi-step execution, although translation may contribute (Appendix~\ref{app:solvability-audit}).

\begin{table}[!h]
\centering
\resizebox{\columnwidth}{!}{%
\begin{tabular}{@{}lrrrrrr@{}}
\toprule
\textbf{Domain} & \textbf{EN} & \textbf{ID} & \textbf{TH} & \textbf{TL} & \textbf{VI} & \textbf{ZH} \\
\midrule
Airline & 92.0\% & 82.0\% & 86.0\% & 84.0\% & 84.0\% & 96.0\% \\
Retail & 95.6\% & 82.5\% & 83.7\% & 85.1\% & 91.2\% & 90.4\% \\
Telecom & 100.0\% & 95.6\% & 96.5\% & 99.1\% & 94.7\% & 100.0\% \\
\bottomrule
\end{tabular}%
}
\caption{Empirical solvability. A task is counted when any of three agent models succeeds in one of three trials. Percentages use 50 Airline tasks and 114 Retail and Telecom tasks each.}
\label{tab:solvability-results}
\end{table}

Beyond solvability, deterministic checks cover 11,416 translation pairs, with targeted model review of 974; separate consistency checks cover 14,890 retail order items and 430 schema literals. Reviewers approved 2,894 corrections across 32 artifacts, after which the reviewed release passed the recorded checks. Each annotation sheet contains a required sample of 90 or 91 units and additional review; at least one native speaker per language completed 167--254 unique units in total. An \texttt{ok} translation is faithful and usable, a \texttt{minor\_issue} affects wording without changing meaning or use, and a \texttt{major\_issue} can change the correct action or result. For each language, we recalibrate the finite-population acceptance number $c$ to its population size and total completed review count. \Cref{tab:main-language-audit} reports all verdicts and the resulting threshold. The observed major-issue count does not exceed $c$ for any language, so all five pass. Appendix~\ref{app:audit-methodology-results} provides the calculation, selection caveat, and full evidence.

\begin{table}[!h]
\centering
\resizebox{\columnwidth}{!}{%
\begin{tabular}{@{}lrrrrrl@{}}
\toprule
\textbf{L2} & \textbf{Reviewed} & \textbf{OK} & \textbf{Minor} & \textbf{Major} & $\boldsymbol{c}$ & \textbf{Decision} \\
\midrule
ID & 254 & 235 & 17 & 2 & 7 & Pass \\
TH & 200 & 189 & 11 & 0 & 5 & Pass \\
TL & 183 & 141 & 42 & 0 & 4 & Pass \\
VI & 167 & 142 & 24 & 1 & 3 & Pass \\
ZH & 200 & 161 & 38 & 1 & 5 & Pass \\
\bottomrule
\end{tabular}%
}
\caption{Native speaker review over all unique units with verdicts. For each language, $c$ is calculated from its population size and reviewed count following the finite-population acceptance sampling procedure; Pass means that the observed major-issue count is at most $c$.}
\label{tab:main-language-audit}
\end{table}

\begin{figure*}[!t]
    \centering
    \includegraphics[width=\linewidth]{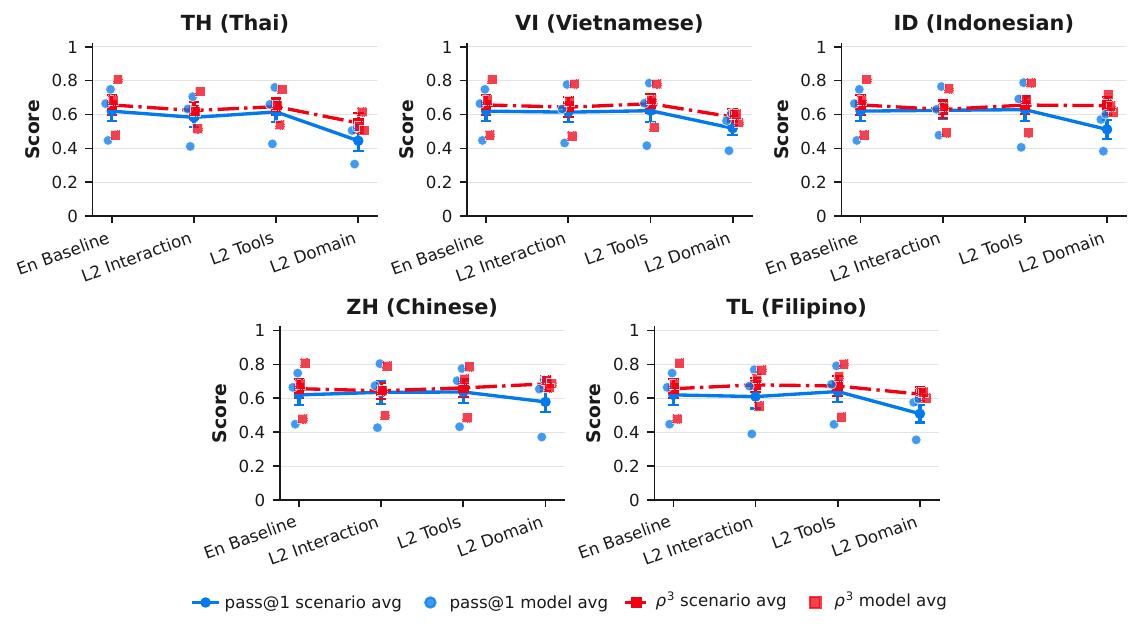}
    \caption{Evaluation results on \seatau. \seatau reveals consistent degradation as the setting becomes less English-centric on all languages in both pass@1 and $\rho^3$ metrics, with most severe drops in low-resource languages with non-Latin scripts.}
    \label{fig:main_result}
\end{figure*}

\section{Experimental Setup} \label{sec:setup-metrics-tasks-params}

\paragraph{Metrics.}
Following \taubench \cite{barres2025tau2}, we report two task metrics. First, pass$@1$ measures quality as the mean success rate of independent trials reaching the expected final state, i.e., pass$@1$=pass$^1$=$\mathbb{E}[r]$. Building on existing metrics, we also derive a robustness measure, $\rho\in[0,1]$, defined as
\begin{equation}
    \rho^q = \frac{\text{pass}^q}{\text{pass@1}}
\end{equation}

where $\rho^q$ denotes the probability that all $q$ independent trials succeed, averaged across tasks. $\rho^q=1$ indicates that the agent performs consistently across multiple runs, while lower $\rho^q$ indicates that the agent may solve a task once but fail to do so reliably across repeated trials.

\begin{figure*}
    \centering
    \includegraphics[width=0.55\linewidth]{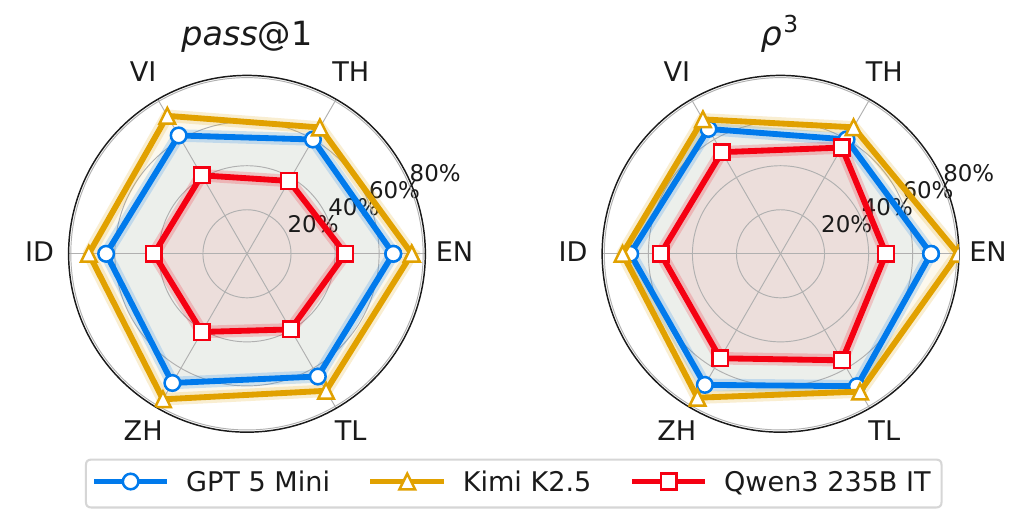}
    \includegraphics[width=0.43\linewidth]{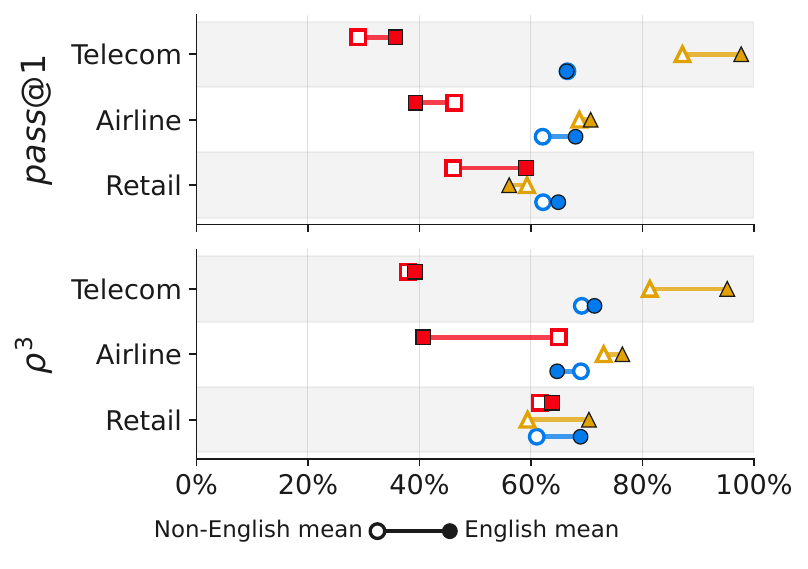}
    \caption{Quality and robustness of varying agent models across languages and domains. \textbf{(left)} All agents yield consistently higher pass$@1$ in English and Chinese indicating a quality bias toward high-resource languages. \textbf{(center)} A similar trend is observed on $\rho^3$, with negative correlation of $\rho^3$ between English and L2 languages especially Thai. \textbf{(right)} pass$@1$ trend holds across most domains and models with some exceptions as described in \Cref{sec:breakdown}.}
    \label{fig:breakdown_result}
\end{figure*}

\paragraph{Task and Domain.} For high-quality evaluation, we use latest versions of retail and airline domains \cite{cuadron2026saber}, which incorporate several corrections and refinements over prior versions \cite{yao2024tau,barres2025tau2}. For the telecom domain, we follow the original \taubench, as its tasks were not updated in the most recent release. We leave exploration of the banking domain~\cite{shi2026tauknowledge} and the voice modality~\cite{ray2026tauvoicebenchmarkingfullduplexvoice} introduced in $\tau^3$-bench to future work.

\paragraph{Hyperparameters and Models.} For all evaluations, we use $q=3$, resulting in the robustness metric $\rho^3$, and Qwen3-235B-A22B-Instruct \cite{qwen3technicalreport} as the user LLM. For the natural-language assertion judge, we use GPT-4.1, following the implementations provided in \taubench. We evaluate three recent LLM agents spanning three model families: GPT-5-Mini \cite{singh2026openaigpt5card} as a proprietary model, Qwen3-235B-A22B-Instruct-2507 as a representative open-source model, and Kimi-K2.5 \cite{kimiteam2026kimik25visualagentic} as a larger-scale open-source model.

For each model, we use the default hyperparameters defined in each provider (see Appendix~\ref{sec:inference-parameters}). We avoid any model-specific hyperparameter tuning to better reflect the out-of-the-box tool-user-agent interaction in the specified scenario and L2.

\section{Results by localization setting}

Across four scenarios, pass$@1$ declines as more of the task environment moves from English to L2, while robustness, model rankings, and domain difficulty vary. We first compare aggregate setting means, then examine model and domain effects.

\subsection{Task quality across settings}
\label{sec:main-pref-degrade}

\paragraph{Average pass$@1$ falls across settings.} Pass$@1$ declines as the benchmark localizes more of the task environment (\Cref{fig:main_result}). The mean falls from $0.62$ in the English Baseline to $0.61$, $0.55$, and $0.51$ in L2 Interaction, L2 Tool, and L2 Domain, respectively. Because L2 Tool includes two agents and the other settings include three, this comparison is descriptive rather than fully matched.

\paragraph{L2 Domain quality varies by language.} Across models and domains, L2 Domain pass$@1$ is lowest for Thai ($0.45$), followed by Filipino ($0.51$), Indonesian ($0.51$), and Vietnamese ($0.52$), while Chinese is highest ($0.58$). Compared with the English Baseline ($0.62$), performance is lower in both Latin- and non-Latin-script languages, although the magnitude differs across languages. These results do not support reducing the gap to script type alone; they instead show language-specific differences in fully localized settings.

\paragraph{Quality and robustness diverge.} $\rho^3$ measures repeated success relative to single-run success. The aggregate $\rho^3$ values are $0.66$, $0.64$, $0.60$, and $0.62$ for the English Baseline, L2 Interaction, L2 Tool, and L2 Domain, respectively. In L2 Domain, Chinese has the highest $\rho^3$ ($0.68$) and Thai the lowest ($0.55$), but several languages retain moderate robustness despite lower pass$@1$. A high $\rho^3$ therefore does not indicate high task quality: it indicates that the tasks an agent solves are solved consistently. Together, the metrics distinguish task success from behavioral reliability.

\subsection{Model and domain rankings shift}
\label{sec:breakdown}

\paragraph{Model rankings vary by setting and metric.} No single agent dominates every condition (\Cref{fig:breakdown_result}). In the English Baseline, Kimi-K2.5 has the highest mean pass$@1$ ($0.75$), followed by GPT-5-mini ($0.66$) and Qwen3-235B-A22B-Instruct-2507 ($0.45$). In L2 Domain, Kimi-K2.5 remains highest on pass$@1$ ($0.61$), but Qwen3-235B has higher mean $\rho^3$ ($0.63$) than GPT-5-mini ($0.62$) and Kimi-K2.5 ($0.59$). The model with the lowest probability of solving a task can therefore still be consistent on the subset of tasks it solves, reinforcing the need to evaluate quality and robustness separately.

\begin{figure*}[!t]
    \centering
    \includegraphics[width=0.9\linewidth]{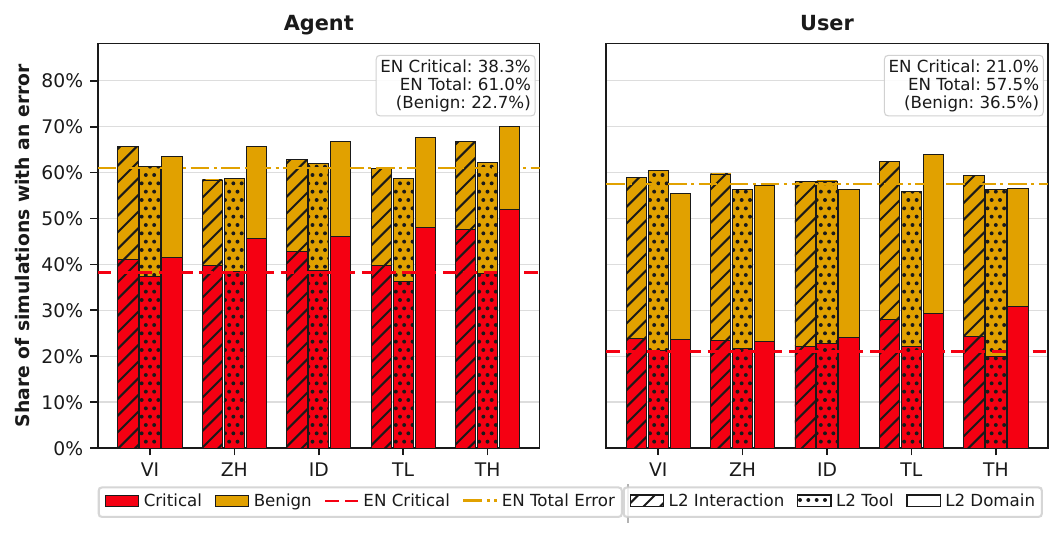}
    \caption{Critical, benign, and correct simulation outcomes for the simulated user and agent. The analysis uses DeepSeek-V4-Flash as the judge, Qwen3-235B-A22B-Instruct-2507 as the user, and GPT-5-mini as the agent.}
    \label{fig:error-scenarios}
\end{figure*}

\paragraph{Domain difficulty varies by setting.} In L2 Domain, Retail has the lowest mean pass$@1$ for all three agents: $0.51$ for GPT-5-mini, $0.48$ for Kimi-K2.5, and $0.25$ for Qwen3-235B. Telecom is easiest for GPT-5-mini and Kimi-K2.5, whereas Airline is easiest for Qwen3-235B. Qwen3-235B shows the clearest reversal across settings: Retail is its strongest English domain ($0.59$) but its weakest L2 Domain ($0.25$). These results show that English domain rankings do not reliably transfer to a fully localized task environment.

\section{Diagnosing the localization gap}

We examine three signals of the localization gap: critical errors, language correctness versus robustness, and cross-language correlations. These analyses distinguish execution failures from patterns of language use and test whether English scores can serve as proxies for L2 performance. Mixed-language tool results and language drift diagnostics appear in Appendices~\ref{app:l2_tool_mix} and~\ref{app:language-use-analysis}.

\subsection{Critical errors in localized settings}
\label{sec:error-analysis}

\begin{figure*}[!t]
    \centering
    \includegraphics[width=\textwidth]{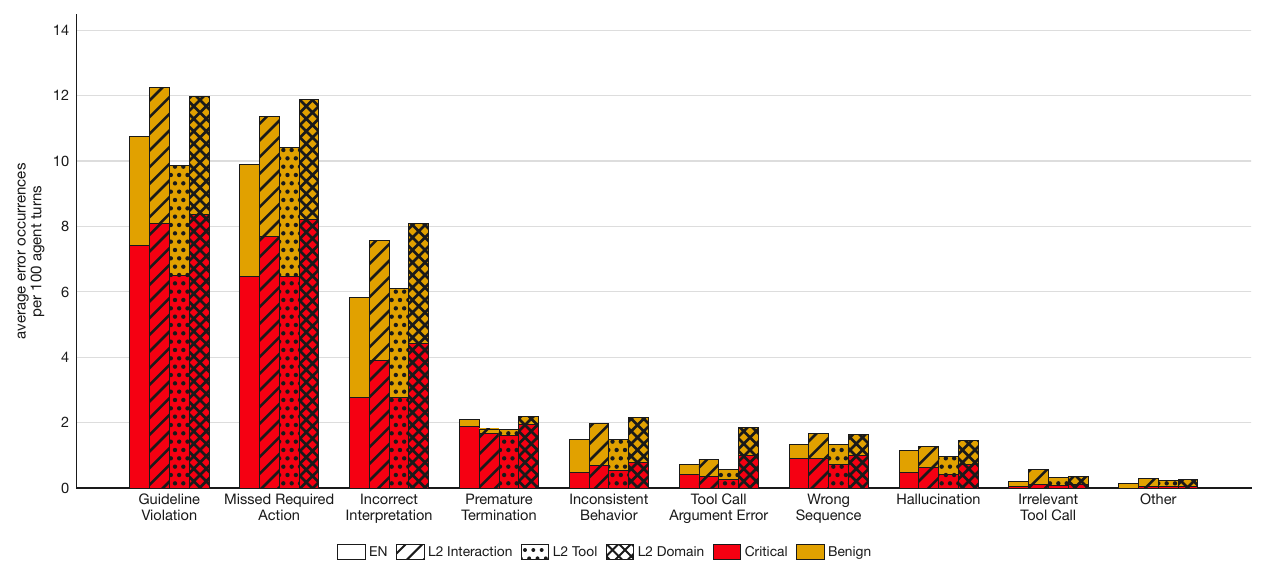}
    \caption{Breakdown of agent errors different across progressive localization scenarios. The trends are consistent across scenarios, with the most errors being Missed Required Action and Guideline Violation. Error taxonomy detailed in Appendix~\ref{app:error-taxonomy}.}
    \label{fig:error_tags_agent}
\end{figure*}

We use the LLM-as-a-Judge framework proposed in \citet{shi2026tauknowledge} to classify errors. Following \citet{barres2025tau2}, a \textit{Critical} error deviates from user intent or is irrecoverable, whereas a \textit{Benign} error does not prevent task completion. For the severity comparison, we consider L2 Interaction and L2 Domain; L2 Tool translates only tool specifications, leaving tool, agent, and user interactions in English. The judge is DeepSeek-V4-Flash, with GPT-5-mini as the agent and Qwen3-235B-A22B-Instruct-2507 as the simulated user.

\paragraph{Agent critical errors rise in L2.} Agent critical error rates are higher in L2 Domain than in the English reference, with Thai showing the largest increase. In L2 Interaction, critical errors range from $39.80\%$ to $47.70\%$ across L2 languages, compared with $38.27\%$ for the English reference. In L2 Domain, they range from $41.54\%$ to $52.06\%$, with Thai at $52.06\%$ and Filipino at $47.98\%$ (\Cref{fig:error-scenarios}). The increase in critical rather than benign errors is consistent with failures in executing the intended task, not only with less fluent dialogue.

\paragraph{User critical errors vary by setting.} User critical error rates are also higher in L2 than in the English reference. They range from $22.20\%$ to $28.10\%$ in L2 Interaction and from $23.10\%$ to $30.80\%$ in L2 Domain, compared with $21.02\%$ for the English reference. These rates show that user-simulation reliability also differs across settings, consistent with prior findings for simulated users in English and dialectal English~\citep{barres2025tau2,seshadri2026lostsimulationllmsimulatedusers}. The error analysis is therefore diagnostic rather than a causal decomposition of the end-to-end performance gap.

\paragraph{Agents omit actions and misread tasks.} \Cref{fig:error_tags_agent} shows three
recurring mistakes across all four scenarios: omitting a required action, violating a task guideline, and misreading the task or user intent. These leading errors occur at roughly $3$ to $6$ occurrences per 100 agent turns, against about $0.7$ to $1.2$ for the next tier and below $0.2$ for the remainder. The ordering is stable across scenarios, but L2 Domain is consistently highest, with differences of about $1.2$ to $1.5$ occurrences per 100 turns relative to the English baseline. Because a single turn may carry several errors and each error several types, these are error type rates rather than a decomposition of failures.

The two tool schema categories are far smaller: \textit{Tool Call Argument Error} reaches $0.91$ and \textit{Tool Call Schema Error} $0.12$ per 100 agent turns under L2 Domain, against $6.3$ and $6.2$ for the two leading policy categories. They rise most steeply in
relative terms from S1 to S4 ($2.8\times$ and $3.8\times$, against $1.3\times$ for the policy errors) yet remain an order of magnitude smaller in absolute terms, and the same ordering holds in S3, where only the schemas are localized. The dominant bottleneck is therefore multi-step policy and domain reasoning rather than reading L2 tool specifications.

\subsection{Language correctness versus robustness}
\label{sec:corr-lang-correctness}

\begin{figure}[!t]
    \centering
    \includegraphics[width=\linewidth]{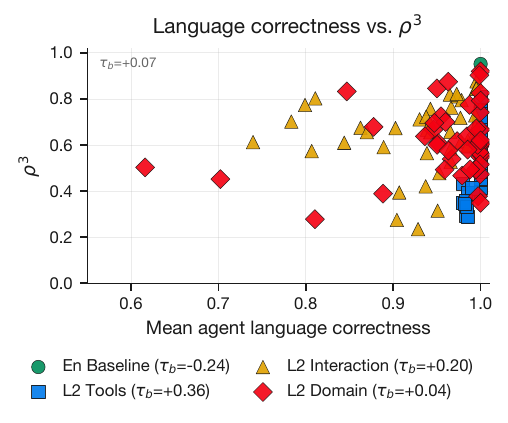}
    \caption{Agent language correctness versus robustness ($\rho^3$) across 168 runs. We report Kendall's rank correlation $\tau_b$ for each scenario, and pooled association. The English Baseline is shown for context, but its language correctness has limited variation with 9 runs.}
    \label{fig:language_correctness_vs_perf}
\end{figure}

Language correctness is the fraction of eligible turns in the expected language; system error trajectories are excluded. Robustness is $\rho^3$, the probability that all three independent trials succeed, averaged across tasks (\Cref{sec:setup-metrics-tasks-params}). We use Kendall's $\tau_b$ to summarize monotonic association because correctness contains tied values (\Cref{fig:language_correctness_vs_perf}).

\paragraph{Correctness--robustness association is weak overall.} Across 168 runs, the pooled Kendall association is weak ($\tau_b=0.07$). The scenario-specific values are $0.36$ in L2 Tool, $0.20$ in L2 Interaction, $0.04$ in L2 Domain, and $-0.24$ in the English Baseline. These values indicate at most a modest monotonic association, not a predictive or causal relationship.

\paragraph{The L2 Tool association is setting specific.} The larger L2 Tool association does not make language correctness a general proxy for robustness. Correctness is concentrated near $1.00$ in L2 Tool, while L2 Domain contains runs with near-perfect correctness and a broad range of $\rho^3$ values. L2 Interaction has more variation in correctness, but its association remains modest. The relationship therefore depends on the scenario and its available predictor variation; correctness alone does not account for the observed robustness differences. Complementary drift diagnostics likewise show that off-target language is measurable and scenario dependent, without establishing that it explains the full performance gap (Appendix~\ref{app:language-use-analysis}).

\subsection{English quality as an L2 proxy}
\label{sec:en-proxy}

\begin{figure}[!t]
    \centering
    \includegraphics[width=\linewidth]{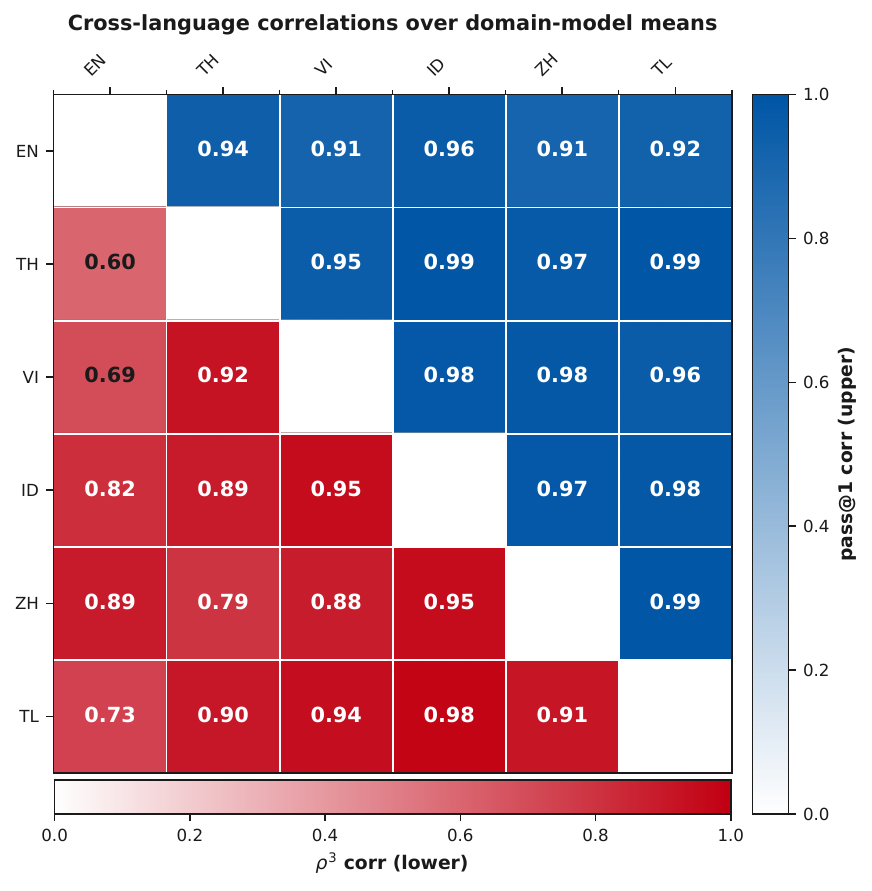}
    \caption{Cross-language Pearson correlations for pass@1 (upper triangle) and $\rho^3$ (lower triangle). Each correlation uses domain--model means after averaging over available scenarios, with three domains and three agent models.}
    \label{fig:metric-corr}
\end{figure}

\paragraph{English pass$@1$ tracks average L2 quality.} English pass$@1$ is a useful coarse predictor of average L2 quality, but it is a poor predictor of repeated run reliability. Because the full benchmark is expensive to run, we correlate language-level pass$@1$ and $\rho^3$ over domain--model means (\Cref{fig:metric-corr}). English correlations with L2 pass$@1$ range from $0.90$ for Vietnamese to $0.96$ for Indonesian, whereas English correlations with L2 $\rho^3$ range from only $0.49$ for Thai to $0.88$ for Chinese.

\paragraph{Filipino correlates across L2 languages.} Filipino is a strong candidate among the L2 languages for a cross-language proxy in this analysis. Its correlations with the other four L2 languages range from $0.94$ to $0.99$ for pass$@1$ and from $0.85$ to $0.94$ for $\rho^3$. A Filipino subset can therefore reduce evaluation cost when a proxy is necessary, but these correlations do not establish validity beyond the five languages, three domains, and three models studied here. For deployment, validate the proxy against the target language.

\section{Conclusion}

% \seatau{} establishes the first agent-focused framework for SEA linguistic diversity, revealing that English agentic capabilities transfer effectively to a superficial L2 interaction but degrade significantly with the inclusion of non-English contexts, due to both agent limitations and reduced user capacity. By exposing fundamental gaps in English-only evaluation methodologies as a proxy of agents capability in SEA languages, \seatau{} provides essential diagnostic tools for sovereign AI development, establishing a rigorous foundation for future research in sovereign agentic AI solutions, ultimately supporting autonomous AI across diverse SEA language communities.

\seatau{} extends agent evaluation beyond translated dialogue by progressively localizing the surfaces on which task completion depends. Its four scenarios separate conversational language from tool schemas and domain context, while the adaptation pipeline preserves a canonical execution layer for comparable scoring. This design makes it possible to ask not only whether an agent can converse in a language, but whether it can reliably complete multi-step tasks when operational information is expressed in that language.

Across five widely spoken SEA languages, performance generally declines as localization reaches deeper into the task environment. The results also show why multilingual agent evaluation should report both quality and robustness. Model rankings shift across settings, and repeated-run consistency can remain moderate even when task success is low. English performance is therefore an incomplete proxy for multilingual readiness. By releasing controlled scenarios, localized assets, and a translation audit methodology, \seatau{} provides a foundation for diagnosing where multilingual agents fail and for evaluating whether future systems improve across languages rather than only on English-centric interfaces.

\section{Limitations}

We evaluate Vietnamese, Indonesian, Thai, Filipino, and Chinese in the retail, airline, and telecom domains. Although the framework can extend to other settings, the findings may not generalize beyond these languages, domains, task distributions, or agent models. Correlations reported over domain--model means are descriptive within this benchmark and should not be interpreted as evidence that one language will remain a reliable proxy in other deployment settings.

The adaptation is also specific to the $\tau$-Bench paradigm \cite{yao2024tau,barres2025tau2,ray2026tauvoicebenchmarkingfullduplexvoice,shi2026tauknowledge}. Its simulator, policies, and canonical tool layer provide experimental control but do not capture every ambiguity, cultural expectation, or interface constraint that arises in deployed systems. Future work should test whether the localization progression transfers to other agent benchmarks and real user environments.

The reported experiments use the immutable experiment snapshot rather than the reviewed release. Passing the native speaker audit and our acceptance thresholds does not imply an error-free translation: the completed reviews contain two major issues in Indonesian and one each in Vietnamese and Chinese. Cross-language gaps may therefore include residual translation error. Future work should rerun the reviewed release and expand probability-based native-speaker review.

\section{Ethical Statement}

We conducted this research with careful attention to ethical considerations throughout the development process. In developing of the translation pipeline and the evaluation framework, we ensure that there is no external sensitive information is exposed throughout the process and the resulting artifacts. We recognize our responsibility to ensure these systems are fair, respectful of different cultures, and beneficial to all users. Our approach focuses on being transparent about what the agents can and cannot do, actively working to identify and address potential biases, and taking steps to prevent misuse. We handle all research data responsibly, protecting privacy while respecting linguistic diversity. This work aims to contribute to AI that serves people equitably across languages and cultures, and we are committed to being accountable for how these technologies impact society.

\section*{Acknowledgments}
This work was supported by the National Research Foundation, Singapore, under its National Large Language Models Funding Initiative, and by the ThaiLLM collaboration through the Digital Economy and Society (DE) Development Fund of the Ministry of Digital Economy and Society, Thailand. Any opinions, findings, conclusions, or recommendations expressed in this material are those of the authors and do not necessarily reflect the views of any of the funding organizations.

We thank Kian Kyars for providing OpenAI API access, Dittaya Wanvarie for providing OpenRouter credits, and SCBX R\&D for providing the Google Cloud Vertex AI resources used in our experiments. We thank Mark Joseph Pasciolco, Manuel Antonio Rufino, and Joseph Imperial for supporting the Filipino language review; and Melanie Bui for Mandarin Chinese language review.

\bibliography{custom}
\clearpage

\appendix
\section{Translation pipeline details}

\subsection{\taubench{} context}
\label{app:tau2-bench-context-for-translation}

Each \taubench{} domain exposes two coupled interfaces. The first contains the benchmark content that the agent sees during an interaction: task definitions, policy and workflow documents, and structured databases. The second contains the executable tool interface, including tool descriptions, schemas, and dynamic response templates.

At runtime, the agent reads the domain context, invokes tools, and communicates with the simulated user. The localization pipeline translates the natural-language surfaces that the agent sees while preserving the canonical values that the execution layer consumes.

\subsection{Translation principles}
\label{app:translation-principles}

The pipeline separates the localized, model-visible surface from the canonical, executable representation. It translates natural-language content, but preserves enum literals, entity identifiers, status codes, and tool argument names in the runtime representation. If these layers are conflated, a translated product name can cause a lookup failure, while an untranslated persona description can break the intended monolingual interaction. The design follows four principles:

\noindent\textbf{Runtime invariance.} The execution layer receives canonical English tokens. Localization changes only the surface shown to the language model.

\noindent\textbf{Terminological consistency.} The same canonical value receives the same target-language realization wherever it appears.

\noindent\textbf{Format fidelity.} The pipeline translates natural-language leaves only. It preserves structural keys, numeric fields, booleans, and program identifiers.

\noindent\textbf{Bidirectional transparency.} The runtime maps localized values to canonical English before tool execution and before metric computation.

The implementation has three components: offline translation materializes language-specific assets, runtime localization adapts the environment presented to the agent, and the audit informs a separately versioned reviewed release.

\subsection{Offline translation}
\label{app:translation-pipeline-offline}

The offline stage uses Gemini 3.1 Flash Lite to translate the static artifacts for each domain. \Cref{tab:artifacts_alignment} summarizes the artifact classes and their roles in runtime alignment.

\begin{table*}[t]
\centering
\resizebox{\linewidth}{!}{
\begin{tabular}{lp{5.2cm}p{6.8cm}}
\hline
    \textbf{Artifact class} & \textbf{Translatable content} & \textbf{Purpose for runtime alignment} \\ \hline
    Task setup & Persona text, instructions, task assertions, and dialogue history & Defines the simulated user's goals and the information needed for evaluation. \\
    Policy and workflow & Document prose & Specifies the agent's allowed behavior and procedures. \\
    Tool descriptions & Descriptions of available tools and their arguments & Guides the agent's interpretation of tool semantics. \\
    Schemas and literals & Descriptions, enum values, and literal alternatives & Defines choices that appear in dynamic tool schemas. \\
    Database content & Natural-language fields, such as names and notes & Supplies textual fields that tool results can expose. \\
    Tool return messages & Exact messages and parameterized templates & Keeps dynamic tool responses in the target language. \\ \hline
    \end{tabular}
}
\caption{Artifact classes translated by the offline pipeline and their roles in runtime alignment.}
\label{tab:artifacts_alignment}
\end{table*}

\paragraph{1. Discover artifacts.} For each selected domain, the pipeline enumerates the benchmark content and executable interface, then assigns each component to the appropriate translation pathway.

\paragraph{2. Extract translatable segments.} The pipeline reduces each component to minimal translation units and applies conservative extraction rules:
\begin{itemize}
    \item Policy documents contribute prose; task records contribute persona text, instructions, assertions, and dialogue history.
    \item Databases contribute only designated natural-language fields, such as names, descriptions, summaries, and notes.
    \item Tool descriptions contribute their natural language explanations, while tool schemas contribute docstring format and values.
    \item Dynamic tool responses contribute messages and parameterized templates.
\end{itemize}

\paragraph{3. Mask canonical tokens.} Before translation, the pipeline identifies values that must remain canonical, including IDs such as \texttt{order\_*} and \texttt{booking\_*}, status values, structural task markers, tool names, and docstring section headers such as \texttt{Args} and \texttt{Returns}. Protection is global for some values and contextual for others. For example, airline literals such as \texttt{basic\_economy} and \texttt{round\_trip} are protected in cabin-class and trip-type contexts, but similar strings in ordinary prose are not. Opaque placeholders temporarily replace protected values and are restored after translation.

\paragraph{4. Translate schema literals first.} The pipeline translates enum labels and \texttt{Literal[...]} values in a dedicated literal mode. It then builds a domain literal map from the localized schema artifacts, including underscore-, space-, and hyphen-separated aliases for each canonical value. This map establishes localized terminology before the pipeline translates longer prose.

\paragraph{5. Translate remaining segments with a glossary.} Task prose, policies, tool docstrings, database leaves, and tool return template are machine translated. The model receives a glossary of schema-derived literals, and the pipeline pre-masks protected values so that localized forms remain consistent. It deduplicates repeated requests when possible and batches segments as structured JSON.

\paragraph{6. Write artifacts and record provenance.} The pipeline writes translated assets while preserving the original structure and invariant values. It records the component, source and target languages, model, translation timestamp, and SHA-256 fingerprints of source files so that a translation can be regenerated and compared with its input.

\subsection{Runtime localization}
\label{app:translation-pipeline-runtime}

Static translation does not cover dynamic tool schemas and tool outputs that the agent sees during inference. The runtime localization layer handles these surfaces using paired source and translated schema artifacts.

\paragraph{7. Build localization maps.} The runtime layer constructs maps for tool descriptions, canonical values and their localized forms, and exact or templated tool responses.

\paragraph{8. Localize the tool schema.} The environment presents localized tool schemas to the agent, including descriptions, choices, default values, and examples when appropriate, while leaving the underlying tool behavior unchanged.

\paragraph{9. Normalize arguments and localize outputs.} Before execution, the runtime layer maps localized arguments supplied by the agent to canonical values. After execution, it localizes structured response payloads and tool return messages so that the agent continues to interact in L2.

\paragraph{10. Canonicalize evaluation payloads.} Before computing metrics, the runtime layer maps localized payloads back to canonical values. The evaluator can therefore compare task outcomes across languages and against the original English benchmark.

\subsection{Translated artifact statistics}
\label{app:translated-artifact-stats}

\Cref{tab:file_coverage,tab:text_volume,tab:schemas,tab:databases} summarize the artifacts produced by the offline translation stage.

\begin{table}[h]
\centering
\begin{tabular}{lcc}
\hline
\textbf{Domain} & \textbf{Files per lang.} & \textbf{Total translated} \\ \hline
Airline & 5 & 25 \\
Retail & 5 & 25 \\
Telecom & 13 & 65 \\ \hline
\textbf{Total} & 23 & 115 \\ \hline
\end{tabular}
\caption{Static artifact files translated per domain. Each domain is translated into five target languages.}
\label{tab:file_coverage}
\end{table}

\begin{table}[h]
\centering
\resizebox{\columnwidth}{!}{%
\begin{tabular}{lcccc}
\hline
\textbf{Domain} & \textbf{Tasks} & \textbf{Docstrings} & \textbf{Policies} & \textbf{Return messages} \\ \hline
Airline & 250 & 14 & 5 & 0 \\
Retail & 570 & 16 & 5 & 0 \\
Telecom & 570 & 43 & 25 & 60 \\ \hline
\end{tabular}%
}
\caption{Translated items by artifact type across the five target languages.}
\label{tab:text_volume}
\end{table}

\begin{table}[h]
\centering
\resizebox{\columnwidth}{!}{%
\begin{tabular}{lccc}
\hline
\textbf{Domain} & \textbf{Total models} & \textbf{Value sets} & \textbf{Localized values} \\ \hline
Airline & 23 & 15 & 21 \\
Retail & 15 & 6 & 14 \\
Telecom & 18 & 12 & 51 \\ \hline
\end{tabular}%
}
\caption{Literals and data models by domain.}
\label{tab:schemas}
\end{table}

\begin{table}[h]
\centering
\resizebox{\columnwidth}{!}{%
\begin{tabular}{lcp{4.5cm}}
\hline
\textbf{Domain} & \textbf{Collections} & \textbf{Record breakdown} \\ \hline
Airline & 3 & flights: 300, users: 500, reservations: 2k \\
Retail & 3 & products: 50, users: 500, orders: 1k \\
Telecom & 6 & plans: 5, devices: 29, lines: 9, customers: 4, bills: 6 \\ \hline
\end{tabular}%
}
\caption{Database artifacts by domain. The pipeline preserves database structure and translates only designated leaf fields.}
\label{tab:databases}
\end{table}

% \subsection{Examples} \label{app:translation-examples} TODO

\section{Audit methodology and results}
\label{app:audit-methodology-results}

The audit tests three possible confounds in multilingual evaluation: whether localization changes empirical task solvability, whether translated artifacts preserve executable content, and whether target-language text remains faithful and usable. We compare English and translated outcomes, check executable content and repeated terms, and ask native speakers to review sampled translation units from the airline, retail, and telecom domains. Approved corrections enter a reviewed release that remains separate from the experiment snapshot. \Cref{tab:audit-overview} summarizes the evidence and its scope.

\begin{table*}[t]
\centering
\resizebox{\linewidth}{!}{%
\begin{tabular}{@{}p{2.6cm}p{5.8cm}p{8.4cm}@{}}
\toprule
\textbf{Stage} & \textbf{Unit and coverage} & \textbf{What it checks} \\
\midrule
Solvability & English and translated task outcomes; 16 discordant cases inspected & Finds translations that can change whether a task is completed successfully. \\
Hallucination and preservation & 11,416 unique source--translation pairs received deterministic checks; 974 received targeted model review and coordinator decisions & Checks names, numbers, identifiers, placeholders, literals, structure, and meaning. \\
Consistency & 14,890 localized order items, 120 product groups, and 430 schema literal checks & Verifies that repeated product names, options, and executable values agree with their canonical forms. \\
Native speaker language review & 167--254 unique reviewed units per language, including a required sample of 90 or 91 units and additional coverage across three domains & Records impact verdicts and error types; the audit-wide threshold is recalibrated to all completed verdicts. \\
Release validation & Corrected artifacts and 2,894 approved corrections across 32 artifacts & Rechecks structure, protected strings, consistency, and data validity while keeping the reported benchmark files unchanged. \\
\bottomrule
\end{tabular}%
}
\caption{Overview of the audit stages used to validate multilingual benchmark artifacts. The middle column gives the reviewed unit and coverage; the final column describes the evidence collected at each stage.}
\label{tab:audit-overview}
\end{table*}

\subsection{Audit stages and scope}
\label{app:audit-scope}

The audit uses a frozen English reference and keeps the experiment snapshot separate from the reviewed release. This separation allows us to correct artifacts for future use without changing the data underlying the reported experiments. Exhaustive deterministic checks suit executable strings because one changed identifier or placeholder can invalidate a tool call; targeted model and human review address meaning and naturalness. The claims below are therefore limited to the checks and samples in \Cref{tab:audit-overview}. We do not claim that every translated sentence received semantic review.

\subsection{Solvability audit}
\label{app:solvability-audit}

As defined in \Cref{sec:l2-translation-audit}, a task is empirically solvable when at least one of nine runs receives reward 1. The task reward is binary because all applicable reward bases must be satisfied. We take the logical OR across the nine runs, so \Cref{tab:solvability-results} measures whether a task succeeds at least once rather than the success rate of an individual run. The denominators are 50 tasks for Airline and 114 each for Retail and Telecom.

Solvability is highest in Telecom: the English rate is 100.0\%, and translated rates range from 94.7\% to 99.1\%. Larger gaps appear in Airline and Retail, particularly for Indonesian, Thai, and Filipino.

We manually examined all 16 translated task instances that were unsolved in L2 but solved in English. Incomplete execution of data refueling occurred in 10 cases, and fixes on the device, such as Wi-Fi Calling or messaging permissions, occurred in 10 cases, with overlap in five cases. In addition, 21 of 144 runs terminated early because of tool errors or step limits.

\paragraph{Execution difficulty dominates the observed gap.} The discordant cases concentrate on tasks where one missed action yields zero reward. The manual review therefore points primarily to compounded execution difficulty under L2 interaction, although it cannot exclude translation as a contributing factor.

\subsection{Hallucination and preservation checks}
\label{app:hallucination-audit}

We treat unsupported additions and changes to executable content as preservation failures. For every unique source--translation pair, deterministic checks compare personal names, numbers, dates, identifiers, placeholders, enum values, tool names, and other runtime literals. They also verify that the translated artifact retains the expected structure. These checks cover all 11,416 pairs.

Targeted review complements the deterministic pass. An independent model reviewer examines 974 pairs selected from high-risk invariant candidates, high-impact semantic strata, code and currency differences, and a low-risk database stratum. The reviewed release records an audit-coordinator decision for all 974 pairs; the first completed human queue contributed 84 decisions, including 63 passes, 11 failures, and 10 deferrals for native judgment. The other 10,442 pairs remain outside semantic review by design.

The reviewed release contains 2,894 approved corrections across 32 artifacts. This number combines corrections identified by several audit stages, so it is not a count of hallucination errors. After correction, the release passed the recorded preservation checks with no correction failures or invariant failures on modified records. We therefore claim exhaustive deterministic coverage and targeted model review with documented coordinator decisions for 974 pairs; we do not claim semantic review of all 11,416 pairs.

\subsection{Consistency audit}
\label{app:consistency-audit}

Consistency validation checks whether the same canonical value is rendered the same way wherever it is exposed to the agent. We compare product names in localized retail orders with the corresponding localized catalog entries, compare product options with their catalog values, and check schema literals in both directions. The bidirectional check catches both multiple translations for one canonical value and collisions in which different canonical values share one translation.

Before correction, the product name comparison found 2,849 mismatched order occurrences across 120 product groups. The counts were 630 for Indonesian, 644 for Thai, 140 for Filipino, 820 for Vietnamese, and 615 for Chinese, as shown in \Cref{tab:consistency-baseline}. The option comparison found zero mismatches. The schema literal population contained 430 checks, or 86 per language, and also found zero issues.

\begin{table}[t]
\centering
\resizebox{\columnwidth}{!}{%
\begin{tabular}{@{}lrr@{}}
\toprule
\textbf{Language} & \textbf{Items checked} & \textbf{Initial name mismatches} \\
\midrule
Indonesian & 2,978 & 630 \\
Thai & 2,978 & 644 \\
Filipino & 2,978 & 140 \\
Vietnamese & 2,978 & 820 \\
Chinese & 2,978 & 615 \\
\midrule
\textbf{Total} & \textbf{14,890} & \textbf{2,849} \\
\bottomrule
\end{tabular}%
}
\caption{Initial consistency findings before approved corrections. Each language contributes the same 2,978 localized order items.}
\label{tab:consistency-baseline}
\end{table}

After correction propagation, the reviewed release has zero remaining product name mismatches, option mismatches, and schema literal issues. We report this result only for these checked value classes; it is not a claim that every natural-language term across every artifact is globally consistent.

\subsection{Native speaker language review}
\label{app:language-audit-method}

We define a translation unit as one unique English source and translated string. Each unit retains its occurrence count and contexts across the airline, retail, and telecom domains. We deduplicate records by stable review-unit identifier when reporting totals; the Indonesian review worksheet contains one repeated, identically reviewed unit. Each review worksheet contains a required sample of 90 or 91 units drawn uniformly without replacement. Additional rows broaden artifact coverage, add a random supplement, or preserve earlier review evidence. For the audit-wide result, we pool all unique completed verdicts and recalibrate the acceptance number to that reviewed count. \Cref{tab:language-verdicts} retains the required-sample--additional split so readers can evaluate both subsets.

\paragraph{Localized values require contextual review.} The L2 Domain interface intentionally translates schema enum labels, defaults, examples, and tool outputs, then maps localized arguments and outputs to canonical values before execution and scoring. A value is therefore not erroneous merely because its displayed L2 form differs from the English canonical form. Reviewers mark a major issue only when a translation changes the intended action or result, or no longer maps to a valid canonical value. Similarly, an English token is marked untranslated only when retaining it is inappropriate in context.

\begin{figure*}[t]
\centering
\includegraphics[width=\textwidth]{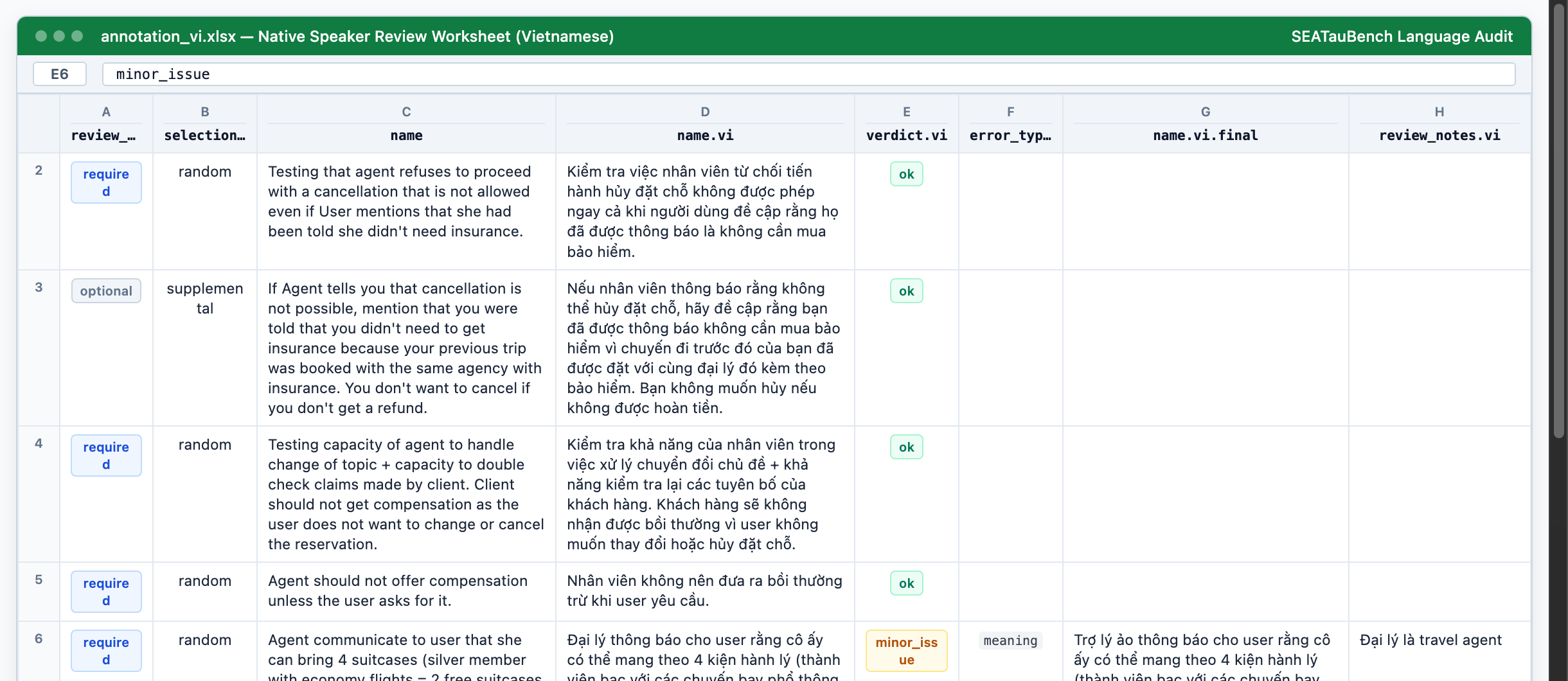}
\caption{Example native speaker review worksheet (\texttt{annotation\_vi.xlsx}, sheet \texttt{airline\_tasks}). Reviewers examine translation units, select impact verdicts and error types, and optionally propose improved translations.}
\label{fig:annotation-sheet-example}
\end{figure*}

Each review worksheet uses three verdicts and one controlled error type field. \Cref{tab:language-schema} summarizes the verdict and error type taxonomy with representative examples. Reviewers can also propose corrected wording and explain their decision. \textbf{These suggestions support adjudication but do not change the experiment snapshot.} \Cref{fig:language-audit-guide} shows the complete instructions provided to native speaker reviewers, and \Cref{fig:annotation-sheet-example} illustrates a representative review worksheet for Vietnamese. The guide describes the prespecified assignment, under which only the required sample was mandatory and additional rows could be skipped. Our audit-wide analysis subsequently reports every completed verdict and applies recalibration.

\begin{table*}[!t]
\centering
\small
\resizebox{\textwidth}{!}{%
\begin{tabular}{@{}lp{6.2cm}p{8.8cm}@{}}
\toprule
\textbf{Category} & \textbf{Interpretation} & \textbf{Representative example} \\
\midrule
\multicolumn{3}{@{}l}{\textbf{Impact verdicts}} \\
\midrule
\texttt{ok} & Meaning, fluency, and executable content are acceptable as written. & A translated sentence preserves its numbers and intent. \\
\texttt{minor\_issue} & A limited meaning, terminology, or language quality problem does not change the task or executable content. & Vietnamese ``agent'' is translated to ``đại lý'' (travel agent). \\
\texttt{major\_issue} & The translation changes meaning or executable content in a way that can affect a task or tool call. & Changing ``gift card'' to ``credit card'' or corrupting \texttt{raj\_sanchez\_7340}. \\
\midrule
\multicolumn{3}{@{}l}{\textbf{Controlled error types}} \\
\midrule
\texttt{meaning} & The translation changes the intended meaning. & Vietnamese ``agent'' rendered as ``đại lý'', commonly associated with a travel agent. \\
\texttt{untranslated} & A term from the source language remains unexpectedly in localized text. & ``user'' remains in an otherwise Chinese sentence. \\
\texttt{inconsistency} & The same canonical value is rendered differently across contexts. & One status label has incompatible L2 forms across equivalent contexts. \\
\texttt{format} & An identifier, placeholder, or markup token is changed. & \texttt{raj\_sanchez\_7340} loses its underscores. \\
\texttt{language\_quality} & The wording is awkward or unnatural, but the task meaning is preserved. & ``kard na debit'' for ``Debit Card''. \\
\texttt{other} & The issue does not fit the controlled categories. & A context-specific reviewer note requiring adjudication. \\
\bottomrule
\end{tabular}%
}
\caption{Language review schema: impact verdicts and controlled error types with representative examples. A verdict records the likely effect on task execution, while the error type describes the nature of the issue.}
\label{tab:language-schema}
\end{table*}

\begin{figure*}[t]
\centering
\begin{minipage}{\textwidth}
\begin{tcolorbox}[
  enhanced,
  width=\textwidth,
  title={Native speaker language audit guide},
  colframe=blue!55!black,
  colbacktitle=blue!12,
  coltitle=black,
  fonttitle=\bfseries,
  boxrule=0.6pt,
  arc=1mm,
  left=2mm,
  right=2mm,
  top=1mm,
  bottom=1mm
]
\footnotesize
\textbf{What to review.} Review every row marked \texttt{required}. Each annotation sheet contains a required sample of 90 or 91 translation units. Rows marked \texttt{optional} broaden coverage to about 200 rows and can be skipped. Under the prespecified reviewer instructions, optional findings did not affect the statistical result.

This audit evaluates the translations used in the completed experiments. It does not change them. A proposed translation is an audit note for a possible future release, not a correction to the experiment data.

\textbf{Review each row.} Compare the English text in \texttt{name} with the translation in \texttt{name.\{lang\}}. A translation is acceptable when it preserves the meaning and facts, gives the agent or user enough information to act correctly, and sounds natural for its purpose.

For each row in the required sample:
\begin{compactenum}
    \item Set \texttt{verdict.\{lang\}} to \texttt{ok}, \texttt{minor\_issue}, or \texttt{major\_issue}.
    \item For an issue, select one error type.
    \item Add a short note when the problem or its direction is not obvious.
    \item Optionally, enter a complete proposed translation in \texttt{name.\{lang\}.final}.
\end{compactenum}
For an \texttt{ok} row, leave the error type and proposed translation blank. Do not mark a translation as an issue only because you prefer different wording.

Edit only \texttt{verdict.\{lang\}}, \texttt{error\_type.\{lang\}}, \texttt{name.\{lang\}.final}, and \texttt{review\_notes.\{lang\}}. Do not add, remove, or reorder rows, columns, or sheets.

\textbf{Choose a verdict.}
\begin{compactitem}
    \item \texttt{ok}: The translation is faithful, understandable, usable, and natural enough for its purpose.
    \item \texttt{minor\_issue}: A genuine language problem exists, but it does not change the meaning or prevent use. Examples include awkward wording, punctuation, or an unsuitable level of formality.
    \item \texttt{major\_issue}: The problem could change the correct action or result. This includes changed meaning, missing or unsupported information, wrong facts, or a damaged value needed to complete a task.
\end{compactitem}

\textbf{Select one error type.}
\begin{compactitem}
    \item \texttt{meaning}: Meaning or content is wrong, missing, or added. In the notes, state whether content is missing or added when that distinction matters.
    \item \texttt{untranslated}: English remains where the target language is needed.
    \item \texttt{inconsistency}: The same entity or term is translated differently without a valid contextual reason.
    \item \texttt{format}: Markdown, a placeholder, code, or layout is damaged.
    \item \texttt{language\_quality}: Wording is unnatural, unclear, or unsuitable in register.
    \item \texttt{other}: Another real problem occurred. Explain it in the notes.
\end{compactitem}
The verdict records the likely effect. The error type describes the problem. A punctuation problem can be minor, while a damaged placeholder can be major.

\textbf{Preserve task values.} People, products, plans, addresses, numbers, dates, times, and amounts must keep the same identity or value. Do not convert currencies or units. Identifiers, placeholders, tool and argument names, email addresses, phone numbers, and structured values must continue to work.

Earlier audits reviewed these values and repeated terminology across all languages. Record a problem if you notice one, but do not repeat those full audits. The annotation sheet's Glossary is a reference, not another review sample.

\textbf{Finish the audit.} Complete every row in the required sample, even after finding a major issue. Before returning the annotation sheet, confirm that every required row has a verdict and that every issue has one error type. Alert the audit coordinator if an issue could change a task result or appears repeatedly.
\end{tcolorbox}
\end{minipage}
\caption{Native speaker language audit guide provided to reviewers.}
\label{fig:language-audit-guide}
\end{figure*}

\paragraph{Finite-population acceptance sampling.} For a population of $N$ unique units and $n$ completed reviews, let $D=\lceil 0.05N\rceil$. We choose the largest integer acceptance number $c$ for which
\[
  P(X \leq c) = \sum_{x=0}^{c}
  \frac{\binom{D}{x}\binom{N-D}{n-x}}{\binom{N}{n}}
  \leq 0.05,
\]
where $X$ is the number of major issues among the reviewed units. Using all completed verdicts yields $c=7$ for Indonesian, $c=5$ for Thai, $c=4$ for Filipino, $c=3$ for Vietnamese, and $c=5$ for Chinese. A language passes when its observed major-issue count is at most its recalculated $c$. Under simple random sampling without replacement, this rule rejects a population with at least 5\% major issues with at least 95\% probability.

\begin{table*}[t]
\centering
\resizebox{\linewidth}{!}{%
\begin{tabular}{@{}lrrrrr@{}}
\toprule
\textbf{Language} & \textbf{Unique units} & \textbf{Required sample} & \textbf{Required complete} & \textbf{Additional selected} & \textbf{Total complete} \\
\midrule
Indonesian & 1,276 & 91 & 91 & 275 & 254 \\
Thai & 1,290 & 90 & 90 & 110 & 200 \\
Filipino & 1,279 & 91 & 91 & 109 & 183 \\
Vietnamese & 1,288 & 90 & 90 & 110 & 167 \\
Chinese & 1,303 & 90 & 90 & 110 & 200 \\
\bottomrule
\end{tabular}%
}
\caption{Native speaker review coverage across the airline, retail, and telecom domains. The required sample is the prespecified sample drawn uniformly without replacement; additional selected includes artifact coverage, supplemental, and earlier review units. Counts deduplicate stable review-unit identifiers.}
\label{tab:language-summary}
\end{table*}

\subsection{Language review results}
\label{app:language-audit-results}

\Cref{tab:language-verdicts} separates the required sample from additional review. Error type counts include every verdict other than \texttt{ok} among completed rows. The audit-wide decision sums completed rows from both groups and compares the total number of major issues with the language-specific $c$ above.

\begin{table*}[t]
\centering
\scriptsize
\resizebox{\linewidth}{!}{%
\begin{tabular}{@{}llrrrrr|rrrrrr@{}}
\toprule
 & & \multicolumn{5}{c|}{\textbf{Verdict}} & \multicolumn{6}{c}{\textbf{Error type}} \\
\textbf{Language} & \textbf{Sample} & $N$ & \textbf{Complete} & \texttt{ok} & \texttt{minor} & \texttt{major} & meaning & untranslated & inconsistency & format & language quality & other \\
\midrule
Indonesian & Required & 91 & 91 & 83 & 8 & 0 & 0 & 6 & 0 & 0 & 2 & 0 \\
 & Additional & 275 & 163 & 152 & 9 & 2 & 1 & 2 & 1 & 1 & 6 & 0 \\
Thai & Required & 90 & 90 & 84 & 6 & 0 & 2 & 0 & 3 & 0 & 1 & 0 \\
 & Additional & 110 & 110 & 105 & 5 & 0 & 3 & 0 & 1 & 0 & 1 & 0 \\
Filipino & Required & 91 & 91 & 67 & 24 & 0 & 3 & 0 & 3 & 1 & 16 & 1 \\
 & Additional & 109 & 92 & 74 & 18 & 0 & 0 & 1 & 4 & 2 & 10 & 1 \\
Vietnamese & Required & 90 & 90 & 79 & 11 & 0 & 4 & 0 & 1 & 0 & 6 & 0 \\
 & Additional & 110 & 77 & 63 & 13 & 1 & 7 & 2 & 0 & 1 & 4 & 0 \\
Chinese & Required & 90 & 90 & 75 & 14 & 1 & 0 & 10 & 2 & 0 & 3 & 0 \\
 & Additional & 110 & 110 & 86 & 24 & 0 & 1 & 20 & 0 & 0 & 3 & 0 \\
\bottomrule
\end{tabular}%
}
\caption{Native speaker verdicts and error types for required and additional units. Complete is the number of unique units with a recorded verdict; error type counts include only completed issue units.}
\label{tab:language-verdicts}
\end{table*}

\paragraph{All five languages meet their acceptance threshold}. Indonesian has two major issues against $c=7$, Thai has zero against $c=5$, Filipino has zero against $c=4$, Vietnamese has one against $c=3$, and Chinese has one against $c=5$.

\paragraph{Issue profiles differ by language.} \textbf{Filipino} has 24 minor issues in its required sample, dominated by language quality, meaning, and consistency. Adjudication accepts ``Mga Batayan ng Domain'' as a usable rendering of ``Domain Basics'' and treats two awkward telecom phrases as minor because they remain actionable; the analogous ``Bill objects'' phrasing in additional review is also minor. Vietnamese has 11 issues in its required sample, mostly meaning and language quality. In two airline task descriptions, ``agent'' becomes ``đại lý'', a term commonly associated with a travel agent. Reviewers marked these cases minor because the instructions remain usable, but the example shows how a role can become narrower in translation. \textbf{Thai} has six minor issues, concentrated in meaning and consistency.

\textbf{Indonesian} has eight minor issues in its required sample: six untranslated and two language-quality findings. Adjudication accepts four previously major findings because the translated status and schema values are intentional L2 labels that map back to canonical values at runtime. A fifth finding is minor because the untranslated word ``available'' does not change the procedure. Additional Indonesian review finds 11 issues, including two major findings: translated payment identifiers and a gift-card/credit-card meaning change.

\textbf{Chinese} has 15 issues in its required sample, including 10 untranslated findings, two consistency findings, three language quality findings, and one major issue associated with localized enum examples in an airline tool description. Across languages, retained English tokens require contextual judgment: some must remain canonical, while others should be translated for natural L2 use.

These patterns motivate the division between deterministic and language review. Identifiers and placeholders must remain resolvable, while localized enum labels and tool values must map unambiguously to their canonical forms. Native review then focuses on meaning, fluency, and terminology after those executable boundaries have been checked.

\subsection{Validation and release}
\label{app:language-validation}

We validate each annotation sheet by checking that every completed row has the expected fields, every verdict other than \texttt{ok} has an error type, and every proposed correction is traceable to one source--translation unit. We then recheck corrected artifacts for structural validity, protected string preservation, and consistency. This process found 2,894 reachable approved corrections, zero correction failures, zero invariant failures on modified records, zero remaining product name mismatches, zero option mismatches, and zero schema literal issues. Data validation passed on the reviewed release.

Corrections remain separate from the experiment snapshot. A correction enters a future benchmark run only when its impact assessment indicates that it can change task behavior, policy interpretation, tool behavior, or the evaluated final state. This rule keeps the audit evidence reproducible while avoiding retrospective changes to the experiments reported in the main paper.

\clearpage
\section{Hyperparameters}
\label{sec:inference-parameters}

\begin{table}[h!]
\centering
\resizebox{\columnwidth}{!}{%
\begin{tabular}{lcc}
\toprule
\textbf{Hyperparameter} & \textbf{Agent LLM} & \textbf{User Sim. LLM} \\
\midrule
Temperature              & 0.0 & 0.0 \\
Nucleus sampling ($p$)   & 1.0 & 1.0 \\
Max.\ generation tokens  & --- & --- \\
\bottomrule
\end{tabular}%
}
\caption{Inference hyperparameters for the L2 scenarios. Values follow the \taubench defaults; parameters not listed are left at the provider model defaults.}
\label{tab:hyperparameters}
\end{table}

All agent LLMs across four scenarios are run with the default inference parameters from \taubench, as shown in \Cref{tab:hyperparameters}.

\section{Compute Budget}
\label{sec:compute-budget}

\begin{table}[h!]
\centering
\resizebox{\columnwidth}{!}{%
\begin{tabular}{lc}
\toprule
\textbf{Scenario} & \textbf{Agent LLM Cost (USD)} \\
\midrule
English Baseline      & \$70.40 \\
L2 Tool    & \$340.73 \\
L2 Interaction        & \$403.27 \\
L2 Domain  & \$440.43 \\
\midrule
\textbf{Total}        & \textbf{\$1,254.83} \\
\bottomrule
\end{tabular}
}
\caption{Approximate agent LLM inference cost (USD) for each experimental scenario, aggregated across three evaluation trials and estimated from token usage with published API pricing.}
\label{tab:compute-agent-cost}
\end{table}

For agent model inference, we use two models across all experimental settings: GPT-5-mini accessed via the OpenRouter API, and Kimi-K2.5 accessed via Azure AI Foundry. \Cref{tab:compute-agent-cost} summarizes the approximate total inference cost per scenario.
For user model inference, we self-host Qwen3-235B-A22B-Instruct-2507 on 8×A100 GPUs (40GB each), resulting in a total of 4,032 GPU hours.

In the translation pipeline, we use Gemini 3.1 Flash Lite for three domains and five target languages. Aggregated across five languages, the pipeline consumes 8.97M input tokens and 7.21M output tokens. The total one-time translation cost for three domains and five languages is approximately \$13.05.

\section{Mixed-language tools cause a modest performance drop}
\label{app:l2_tool_mix}

\begin{figure}[ht]
\centering
\includegraphics[width=\linewidth]{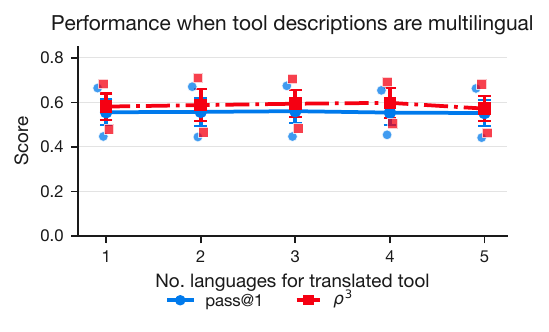}
\caption{Task quality and robustness as the number of languages used in tool specifications increases from one (English) to five. GPT-5-mini and Qwen3-235B-A22B-Instruct show a small initial drop when a second language is added, followed by a plateau.}
\label{fig:perf_tool_mix}
\end{figure}

We extend the L2 Tool experiment to measure whether agents remain effective when different tools use different languages. Mix-1 uses English for every tool. Mix-2 adds Thai, Mix-3 adds Vietnamese, Mix-4 adds Indonesian, and Mix-5 adds Chinese. The language assigned to each tool remains fixed throughout a run; for example, a \texttt{get\_item} tool assigned to Thai remains in Thai for every example. Each added non-English language localizes three tools, while the remaining tools stay in English. Thus the number of languages and the fraction of English tool definitions change across mixtures.

\Cref{fig:perf_tool_mix} shows a modest overall effect. Mean pass$@1$ decreases from $0.68$ with English-only tools to $0.55$ in the most multilingual mixture, a difference of $0.13$. The corresponding decrease in $\rho^3$ is $0.11$. The curve is relatively flat after the first additional language: increasing the number of languages beyond Mix-2 does not produce another comparably large decline in this evaluation. Mixed-language tool specifications therefore reduce performance in this setting, but the result does not indicate a proportional loss as more languages are added.
\section{Agent language use and drift analysis}
\label{app:language-use-analysis}

\begin{figure}
    \centering
    \includegraphics[width=\linewidth]{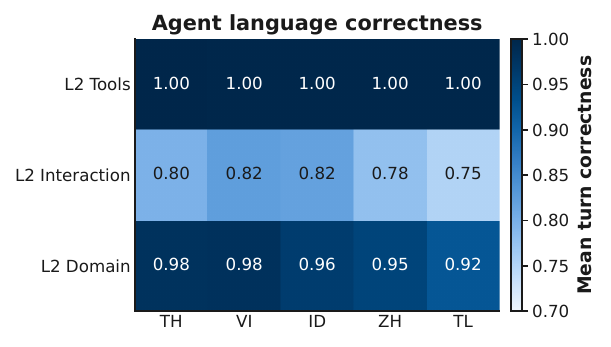}
    \caption{Mean agent language correctness for each turn by setting and target language. The expected language is English in the English Baseline and L2 Tool settings, and the target language in L2 Interaction and L2 Domain. Correctness is the fraction of eligible turns for which fastText detects the expected language.}
    \label{fig:language-correctness-heatmap}
\end{figure}

\begin{figure}
    \centering
    \includegraphics[width=\linewidth]{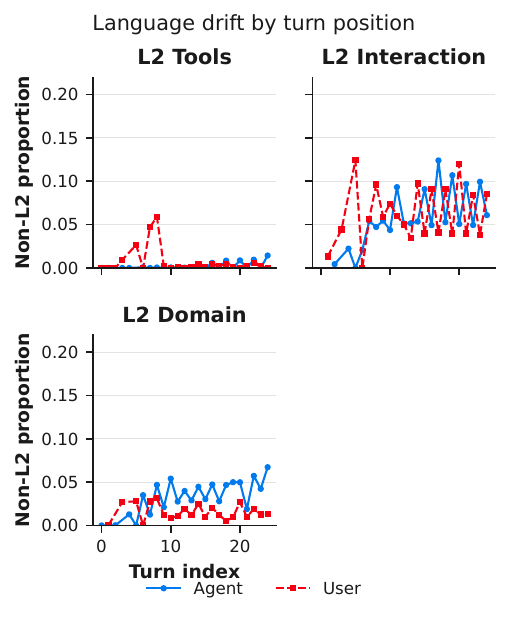}
    \caption{Proportion of turns that do not match the setting's expected language, by turn position, setting, and speaker role. The roles are the agent and the simulated user.}
    \label{fig:language_drift_by_turn_position}
\end{figure}

\begin{figure}[!t]
    \centering
    \includegraphics[width=\linewidth]{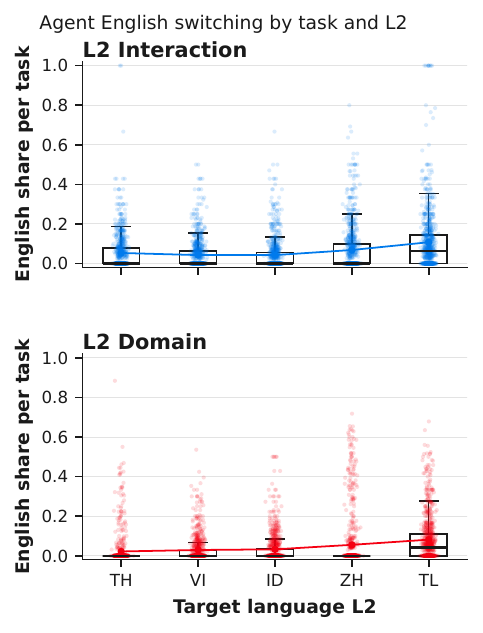}
    \caption{Per-task share of agent turns emitted in English, by target language, in the two settings with L2 dialogue. Each dot represents a task, and the overlaid line shows the per-language mean.}
    \label{fig:agent_english_share_boxplots}
\end{figure}

We measure agent language behavior at three levels: correctness for each turn, the position of off-target turns in a conversation, and the share of agent turns that are in English. For each eligible turn, fastText identifies the output language, which we compare with the language expected by the setting. We report the mean correctness in \Cref{fig:language-correctness-heatmap}, the position of off-target turns in \Cref{fig:language_drift_by_turn_position}, and English output shares in \Cref{fig:agent_english_share_boxplots}.

\paragraph{Correctness for individual turns remains high except during L2 interaction.}
In L2 Tool, mean correctness is approximately $1.00$ for all five target languages. Because the dialogue remains in English in this setting, translating tool schemas alone does not induce measurable language drift. L2 Domain also maintains high correctness, ranging from $0.92$ to $0.98$: Thai and Vietnamese reach $0.98$, while Filipino reaches $0.92$. Correctness is lower in L2 Interaction, ranging from $0.75$ to $0.82$, with the lowest values for Filipino ($0.75$) and Chinese ($0.78$). The difference between settings exceeds the differences among languages within a setting. Moving from L2 Domain to L2 Interaction reduces correctness by approximately $0.15$--$0.20$ for each language, whereas the within-setting spread is at most $\sim 0.07$. The interaction regime therefore explains more of the observed drift than language identity alone.

\paragraph{Off-target output becomes more common later in the conversation.}
\Cref{fig:language_drift_by_turn_position} shows how off-target output changes with turn position. In L2 Tool, the off-target proportion remains negligible for both roles, consistent with the English dialogue specified by that setting. In L2 Interaction and L2 Domain, the agent's off-target share increases with turn index; the Pearson correlations between turn position and off-target proportion are approximately $0.65$ and $0.73$, respectively. This pattern is consistent with accumulation across turns rather than a constant error rate per turn. The two settings differ in the roles that contribute to the increase. In L2 Interaction, both the simulated user and the agent produce off-target turns, and the agent share reaches approximately $0.12$ in later turns. In L2 Domain, the simulated user's share remains approximately $0.015$, while the agent's share rises to approximately $0.07$ in later turns. The co-occurrence of off-target output from both roles in L2 Interaction is consistent with the lower correctness for individual turns in that setting.

\paragraph{Off-target agent output is usually English.}

\Cref{fig:agent_english_share_boxplots} shows that English accounts for most off-target agent output, although English switching remains uncommon for most languages. For Thai, Vietnamese, Indonesian, and Chinese, the median share of English agent turns for each task is $0$ in both settings; the right-skewed distributions have means of only a few percent. Filipino is the exception. Its English share has the highest mean and median in both settings: $0.11$ and $0.06$ in L2 Interaction, and $0.08$ and $0.04$ in L2 Domain, respectively. The right-skewed distributions indicate that switching is concentrated in a minority of tasks, rather than spread uniformly across tasks.

\paragraph{Language drift is measurable but does not explain most performance variation.}
The three analyses identify drift as scenario-dependent: it is strongest in L2 Interaction, concentrated in Filipino, increases later in conversations, and appears mainly as English switching. Drift is therefore a secondary symptom of the harder crosslingual settings, not a sufficient explanation for their task-performance gaps. Across all scenarios, language correctness has a weak pooled association with $\rho^3$ (Pearson $r=0.10$, $R^2=0.009$, $n=153$). L2 Tool is the only setting with a strong positive correlation, but correctness is saturated near $1.00$ there and provides little usable variation. The correlations in L2 Interaction and L2 Domain are weak ($|r|\lesssim 0.25$). These observational correlations do not establish that language drift causes the differences in $\rho^3$.
\section{All results}
% Required packages in the ACL preamble:
% \usepackage{booktabs}
% \usepackage{graphicx}
% These portrait tables use table* to span both ACL columns.
% Columns that were entirely N/A or blank in the source CSV after model filtering are omitted.
% Scores are formatted to three decimals, matching the earlier ACL table style.
\begin{table*}[t]
\centering
\small
\setlength{\tabcolsep}{4pt}
\label{tab:sea-taubench-s1}
\begin{tabular}{@{}llrrrr@{}}
\toprule
Domain & Model & \multicolumn{4}{c}{EN} \\
\cmidrule(lr){3-6}
 &  & p@1 & $p^2$ & $p^3$ & $\rho^3$ \\
\midrule
Airline & GPT-5-mini & 0.693 & 0.593 & \textbf{0.540} & \textbf{0.779} \\
 & Qwen3-235B & 0.500 & 0.387 & 0.340 & 0.680 \\
 & Kimi-K2.5 & \textbf{0.707} & \textbf{0.600} & \textbf{0.540} & 0.764 \\
\midrule
Retail & GPT-5-mini & \textbf{0.655} & \textbf{0.497} & \textbf{0.412} & 0.629 \\
 & Qwen3-235B & 0.579 & 0.415 & 0.342 & 0.591 \\
 & Kimi-K2.5 & 0.561 & 0.442 & 0.395 & \textbf{0.704} \\
\midrule
Telecom & GPT-5-mini & 0.713 & 0.588 & 0.509 & 0.714 \\
 & Qwen3-235B & 0.357 & 0.205 & 0.132 & 0.370 \\
 & Kimi-K2.5 & \textbf{0.997} & \textbf{0.953} & \textbf{0.930} & \textbf{0.933} \\
\bottomrule
\end{tabular}%
\par\vspace{1mm}
\begin{minipage}{\textwidth}
\caption{English Baseline (S1) results. Where EN denotes English, p@1 denotes pass@1, $p^2$ denotes pass$^2$, and $p^3$ denotes pass$^3$. Bold values indicate the best score within each domain--metric group. Qwen3-235B refers to \texttt{Qwen3-235B-A22B-Instruct-2507}.}
\end{minipage}
\end{table*}

\begin{table*}[t]
\centering
\scriptsize
\setlength{\tabcolsep}{2.2pt}
\label{tab:sea-taubench-s3}
\resizebox{\textwidth}{!}{%
\begin{tabular}{@{}llrrrrrrrrrrrrrrrrrrrr@{}}
\toprule
Domain & Model & \multicolumn{4}{c}{VI} & \multicolumn{4}{c}{TH} & \multicolumn{4}{c}{ID} & \multicolumn{4}{c}{ZH} & \multicolumn{4}{c}{TL} \\
\cmidrule(lr){3-6} \cmidrule(lr){7-10} \cmidrule(lr){11-14} \cmidrule(lr){15-18} \cmidrule(lr){19-22}
 &  & p@1 & $p^2$ & $p^3$ & $\rho^3$ & p@1 & $p^2$ & $p^3$ & $\rho^3$ & p@1 & $p^2$ & $p^3$ & $\rho^3$ & p@1 & $p^2$ & $p^3$ & $\rho^3$ & p@1 & $p^2$ & $p^3$ & $\rho^3$ \\
\midrule
Airline & GPT-5-mini & 0.640 & 0.500 & 0.420 & 0.656 & 0.593 & 0.473 & 0.400 & 0.674 & 0.627 & 0.447 & 0.360 & 0.574 & 0.620 & 0.460 & 0.380 & 0.613 & 0.620 & 0.527 & 0.480 & 0.774 \\
 & Qwen3-235B & 0.480 & 0.347 & 0.280 & 0.583 & 0.487 & 0.373 & 0.320 & 0.657 & 0.493 & 0.347 & 0.260 & 0.527 & 0.460 & 0.300 & 0.220 & 0.478 & 0.440 & 0.313 & 0.260 & 0.591 \\
 & Kimi-K2.5 & \textbf{0.733} & \textbf{0.660} & \textbf{0.600} & \textbf{0.818} & \textbf{0.687} & \textbf{0.580} & \textbf{0.520} & \textbf{0.757} & \textbf{0.740} & \textbf{0.653} & \textbf{0.600} & \textbf{0.811} & \textbf{0.760} & \textbf{0.640} & \textbf{0.580} & \textbf{0.763} & \textbf{0.747} & \textbf{0.653} & \textbf{0.600} & \textbf{0.804} \\
\midrule
Retail & GPT-5-mini & \textbf{0.696} & \textbf{0.570} & \textbf{0.500} & \textbf{0.718} & \textbf{0.699} & \textbf{0.538} & \textbf{0.430} & 0.615 & \textbf{0.684} & \textbf{0.547} & \textbf{0.456} & \textbf{0.667} & 0.675 & 0.506 & 0.412 & 0.610 & 0.655 & 0.518 & 0.430 & 0.656 \\
 & Qwen3-235B & 0.556 & 0.418 & 0.333 & 0.600 & 0.526 & 0.383 & 0.325 & 0.617 & 0.608 & 0.462 & 0.386 & 0.635 & 0.591 & 0.442 & 0.351 & 0.594 & 0.573 & 0.444 & 0.386 & 0.674 \\
 & Kimi-K2.5 & 0.687 & 0.550 & 0.465 & 0.677 & 0.573 & 0.430 & 0.360 & \textbf{0.627} & 0.640 & 0.483 & 0.386 & 0.603 & \textbf{0.722} & \textbf{0.597} & \textbf{0.526} & \textbf{0.729} & \textbf{0.675} & \textbf{0.541} & \textbf{0.474} & \textbf{0.701} \\
\midrule
Telecom & GPT-5-mini & 0.561 & 0.430 & 0.368 & 0.656 & 0.605 & 0.436 & 0.342 & 0.565 & 0.582 & 0.453 & 0.404 & 0.693 & 0.728 & 0.608 & 0.518 & 0.711 & 0.740 & 0.617 & 0.535 & 0.723 \\
 & Qwen3-235B & 0.260 & 0.129 & 0.061 & 0.236 & 0.222 & 0.108 & 0.061 & 0.276 & 0.333 & 0.170 & 0.105 & 0.316 & 0.228 & 0.132 & 0.097 & 0.423 & 0.155 & 0.088 & 0.061 & 0.396 \\
 & Kimi-K2.5 & \textbf{0.912} & \textbf{0.836} & \textbf{0.772} & \textbf{0.846} & \textbf{0.854} & \textbf{0.763} & \textbf{0.702} & \textbf{0.822} & \textbf{0.915} & \textbf{0.839} & \textbf{0.772} & \textbf{0.843} & \textbf{0.930} & \textbf{0.868} & \textbf{0.816} & \textbf{0.877} & \textbf{0.883} & \textbf{0.784} & \textbf{0.702} & \textbf{0.795} \\
\bottomrule
\end{tabular}%
}
\par\vspace{1mm}
\begin{minipage}{\textwidth}
\caption{L2 Interaction (S2) results across Vietnamese (VI), Thai (TH), Indonesian (ID), Chinese (ZH), and Filipino (TL). Here, p@1 denotes pass@1, $p^2$ denotes pass$^2$, and $p^3$ denotes pass$^3$. Bold values indicate the best score within each domain--metric group. Qwen3-235B refers to \texttt{Qwen3-235B-A22B-Instruct-2507}.}
\end{minipage}
\end{table*}

\begin{table*}[t]
\centering
\scriptsize
\setlength{\tabcolsep}{2.2pt}
\label{tab:sea-taubench-s2-mono}
\resizebox{\textwidth}{!}{%
\begin{tabular}{@{}llrrrrrrrrrrrrrrrrrrrr@{}}
\toprule
Domain & Model & \multicolumn{4}{c}{VI} & \multicolumn{4}{c}{TH} & \multicolumn{4}{c}{ID} & \multicolumn{4}{c}{ZH} & \multicolumn{4}{c}{TL} \\
\cmidrule(lr){3-6} \cmidrule(lr){7-10} \cmidrule(lr){11-14} \cmidrule(lr){15-18} \cmidrule(lr){19-22}
 &  & p@1 & $p^2$ & $p^3$ & $\rho^3$ & p@1 & $p^2$ & $p^3$ & $\rho^3$ & p@1 & $p^2$ & $p^3$ & $\rho^3$ & p@1 & $p^2$ & $p^3$ & $\rho^3$ & p@1 & $p^2$ & $p^3$ & $\rho^3$ \\
\midrule
Airline & GPT-5-mini
 & 0.653 & 0.527 & 0.460 & 0.704
 & 0.620 & 0.473 & 0.400 & 0.645
 & 0.687 & 0.567 & 0.480 & 0.699
 & 0.687 & \textbf{0.593} & 0.520 & 0.757
 & 0.667 & 0.580 & 0.540 & 0.810 \\
 & Kimi-K2.5
 & \textbf{0.727} & \textbf{0.620} & \textbf{0.560} & \textbf{0.770}
 & \textbf{0.713} & \textbf{0.587} & \textbf{0.500} & \textbf{0.701}
 & \textbf{0.753} & \textbf{0.640} & \textbf{0.560} & \textbf{0.744}
 & \textbf{0.700} & \textbf{0.593} & \textbf{0.540} & \textbf{0.771}
 & \textbf{0.773} & \textbf{0.693} & \textbf{0.640} & \textbf{0.828} \\
 & Qwen3-235B
 & 0.447 & 0.313 & 0.260 & 0.582
 & 0.380 & 0.253 & 0.220 & 0.579
 & 0.400 & 0.233 & 0.180 & 0.450
 & 0.480 & 0.313 & 0.200 & 0.417
 & 0.467 & 0.320 & 0.240 & 0.514 \\
\midrule
Retail & GPT-5-mini
 & \textbf{0.681} & \textbf{0.532} & \textbf{0.430} & 0.631
 & \textbf{0.646} & \textbf{0.477} & \textbf{0.377} & 0.584
 & \textbf{0.675} & \textbf{0.526} & 0.430 & 0.637
 & \textbf{0.719} & \textbf{0.579} & \textbf{0.482} & 0.670
 & \textbf{0.661} & \textbf{0.515} & \textbf{0.439} & \textbf{0.664} \\
 & Kimi-K2.5
 & 0.658 & 0.497 & 0.404 & 0.614
 & 0.602 & 0.456 & 0.360 & 0.598
 & 0.658 & 0.523 & \textbf{0.439} & \textbf{0.667}
 & 0.673 & 0.532 & 0.456 & \textbf{0.678}
 & 0.629 & 0.482 & 0.395 & 0.628 \\
 & Qwen3-235B
 & 0.547 & 0.412 & 0.351 & \textbf{0.642}
 & 0.561 & 0.421 & 0.351 & \textbf{0.626}
 & 0.550 & 0.418 & 0.351 & 0.638
 & 0.594 & 0.433 & 0.360 & 0.606
 & 0.570 & 0.421 & 0.342 & 0.600 \\
\midrule
Telecom & GPT-5-mini
 & 0.664 & 0.550 & 0.482 & 0.726
 & 0.713 & 0.588 & 0.518 & 0.727
 & 0.713 & 0.596 & 0.518 & 0.727
 & 0.705 & 0.576 & 0.500 & 0.709
 & 0.719 & 0.594 & 0.509 & 0.708 \\
 & Kimi-K2.5
 & \textbf{0.971} & \textbf{0.947} & \textbf{0.930} & \textbf{0.958}
 & \textbf{0.965} & \textbf{0.936} & \textbf{0.912} & \textbf{0.945}
 & \textbf{0.953} & \textbf{0.927} & \textbf{0.904} & \textbf{0.949}
 & \textbf{0.953} & \textbf{0.909} & \textbf{0.868} & \textbf{0.911}
 & \textbf{0.971} & \textbf{0.944} & \textbf{0.921} & \textbf{0.949} \\
 & Qwen3-235B
 & 0.254 & 0.135 & 0.088 & 0.346
 & 0.339 & 0.208 & 0.140 & 0.413
 & 0.269 & 0.155 & 0.105 & 0.390
 & 0.222 & 0.135 & 0.096 & 0.432
 & 0.301 & 0.167 & 0.105 & 0.349 \\
\bottomrule
\end{tabular}%
}
\par\vspace{1mm}
\begin{minipage}{\textwidth}
\caption{L2 Tool (S3) results in monolingual settings across Vietnamese (VI), Thai (TH), Indonesian (ID), Chinese (ZH), and Filipino (TL). Here, p@1 denotes pass@1, $p^2$ denotes pass$^2$, and $p^3$ denotes pass$^3$. Bold values indicate the best score within each domain--metric group. Qwen3-235B refers to \texttt{Qwen3-235B-A22B-Instruct-2507}.}
\end{minipage}
\end{table*}

\begin{table*}[t]
\centering
\scriptsize
\setlength{\tabcolsep}{2.2pt}
\label{tab:sea-taubench-s2-multi}
\resizebox{\textwidth}{!}{%
\begin{tabular}{@{}llrrrrrrrrrrrrrrrr@{}}
\toprule
Domain & Model & \multicolumn{4}{c}{Bi} & \multicolumn{4}{c}{Tri} & \multicolumn{4}{c}{Quad} & \multicolumn{4}{c}{Multi} \\
\cmidrule(lr){3-6} \cmidrule(lr){7-10} \cmidrule(lr){11-14} \cmidrule(lr){15-18}
 &  & p@1 & $p^2$ & $p^3$ & $\rho^3$ & p@1 & $p^2$ & $p^3$ & $\rho^3$ & p@1 & $p^2$ & $p^3$ & $\rho^3$ & p@1 & $p^2$ & $p^3$ & $\rho^3$ \\
\midrule
Airline & GPT-5-mini & \textbf{0.680} & \textbf{0.560} & \textbf{0.480} & \textbf{0.706} & \textbf{0.660} & \textbf{0.560} & \textbf{0.480} & \textbf{0.727} & \textbf{0.640} & \textbf{0.520} & \textbf{0.480} & \textbf{0.750} & \textbf{0.640} & \textbf{0.513} & \textbf{0.440} & \textbf{0.688} \\
 & Qwen3-235B & 0.440 & 0.280 & 0.200 & 0.455 & 0.413 & 0.267 & 0.200 & 0.484 & 0.413 & 0.287 & 0.240 & 0.581 & 0.413 & 0.267 & 0.220 & 0.533 \\
\midrule
Retail & GPT-5-mini & \textbf{0.667} & \textbf{0.541} & \textbf{0.465} & \textbf{0.697} & \textbf{0.667} & \textbf{0.538} & \textbf{0.456} & \textbf{0.684} & \textbf{0.652} & \textbf{0.485} & \textbf{0.386} & 0.592 & \textbf{0.696} & \textbf{0.547} & \textbf{0.447} & \textbf{0.642} \\
 & Qwen3-235B & 0.596 & 0.456 & 0.386 & 0.648 & 0.535 & 0.398 & 0.325 & 0.607 & 0.585 & 0.447 & 0.377 & \textbf{0.644} & 0.570 & 0.401 & 0.298 & 0.523 \\
\midrule
Telecom & GPT-5-mini & \textbf{0.664} & \textbf{0.541} & \textbf{0.482} & \textbf{0.726} & \textbf{0.696} & \textbf{0.558} & \textbf{0.491} & \textbf{0.705} & \textbf{0.670} & \textbf{0.550} & \textbf{0.491} & \textbf{0.733} & \textbf{0.652} & \textbf{0.520} & \textbf{0.465} & \textbf{0.713} \\
 & Qwen3-235B & 0.298 & 0.155 & 0.088 & 0.295 & 0.392 & 0.213 & 0.140 & 0.357 & 0.365 & 0.181 & 0.105 & 0.288 & 0.342 & 0.181 & 0.114 & 0.333 \\
\bottomrule
\end{tabular}%
}
\par\vspace{1mm}
\begin{minipage}{\textwidth}
\caption{L2 Tool (S3) results for multilingual settings, including bilingual (Bi), trilingual (Tri), quadlingual (Quad), and multilingual (Multi) configurations. Here, p@1 denotes pass@1, $p^2$ denotes pass$^2$, and $p^3$ denotes pass$^3$. Bold values indicate the best score within each domain--metric group. Qwen3-235B refers to \texttt{Qwen3-235B-A22B-Instruct-2507}.}
\end{minipage}
\end{table*}

\begin{table*}[h]
\centering
\scriptsize
\setlength{\tabcolsep}{2.2pt}
\label{tab:sea-taubench-s4}
\resizebox{\textwidth}{!}{%
\begin{tabular}{@{}llrrrrrrrrrrrrrrrrrrrr@{}}
\toprule
Domain & Model & \multicolumn{4}{c}{VI} & \multicolumn{4}{c}{TH} & \multicolumn{4}{c}{ID} & \multicolumn{4}{c}{ZH} & \multicolumn{4}{c}{TL} \\
\cmidrule(lr){3-6} \cmidrule(lr){7-10} \cmidrule(lr){11-14} \cmidrule(lr){15-18} \cmidrule(lr){19-22}
 &  & p@1 & $p^2$ & $p^3$ & $\rho^3$ & p@1 & $p^2$ & $p^3$ & $\rho^3$ & p@1 & $p^2$ & $p^3$ & $\rho^3$ & p@1 & $p^2$ & $p^3$ & $\rho^3$ & p@1 & $p^2$ & $p^3$ & $\rho^3$ \\
\midrule
Airline & GPT-5-mini & 0.540 & 0.400 & 0.320 & 0.593 & \textbf{0.593} & \textbf{0.453} & 0.360 & 0.607 & \textbf{0.600} & \textbf{0.540} & \textbf{0.500} & 0.833 & 0.607 & 0.493 & 0.440 & 0.725 & 0.560 & 0.433 & 0.380 & 0.679 \\
 & Qwen3-235B & 0.487 & \textbf{0.460} & \textbf{0.440} & \textbf{0.903} & 0.460 & 0.413 & \textbf{0.380} & \textbf{0.826} & 0.480 & 0.440 & 0.420 & \textbf{0.875} & 0.500 & 0.473 & \textbf{0.460} & \textbf{0.920} & 0.473 & 0.420 & \textbf{0.400} & \textbf{0.846} \\
 & Kimi-K2.5 & \textbf{0.560} & 0.427 & 0.340 & 0.607 & 0.547 & 0.387 & 0.300 & 0.548 & \textbf{0.600} & 0.487 & 0.420 & 0.700 & \textbf{0.660} & \textbf{0.507} & 0.440 & 0.667 & \textbf{0.600} & \textbf{0.487} & \textbf{0.400} & 0.667 \\
\midrule
Retail & GPT-5-mini & \textbf{0.585} & \textbf{0.427} & \textbf{0.333} & \textbf{0.569} & 0.325 & 0.175 & 0.123 & 0.378 & \textbf{0.541} & \textbf{0.392} & \textbf{0.325} & \textbf{0.601} & 0.596 & \textbf{0.450} & \textbf{0.368} & \textbf{0.617} & \textbf{0.488} & \textbf{0.333} & \textbf{0.263} & 0.539 \\
 & Qwen3-235B & 0.289 & 0.205 & 0.149 & 0.516 & 0.178 & 0.143 & \textbf{0.132} & \textbf{0.742} & 0.243 & 0.170 & 0.140 & 0.576 & 0.266 & 0.181 & 0.149 & 0.560 & 0.260 & 0.187 & 0.158 & \textbf{0.608} \\
 & Kimi-K2.5 & 0.567 & 0.392 & 0.268 & 0.473 & \textbf{0.327} & \textbf{0.190} & 0.114 & 0.349 & 0.433 & 0.287 & 0.202 & 0.467 & \textbf{0.607} & \textbf{0.450} & 0.366 & 0.603 & 0.444 & 0.292 & 0.219 & 0.493 \\
\midrule
Telecom & GPT-5-mini & 0.564 & 0.427 & 0.360 & \textbf{0.638} & 0.599 & 0.447 & 0.360 & 0.601 & 0.497 & 0.325 & 0.246 & 0.495 & 0.760 & 0.652 & 0.588 & 0.774 & 0.681 & 0.547 & \textbf{0.474} & \textbf{0.696} \\
 & Qwen3-235B & 0.383 & 0.213 & 0.149 & 0.389 & 0.284 & 0.143 & 0.079 & 0.278 & 0.427 & 0.295 & 0.211 & 0.494 & 0.330 & 0.205 & 0.149 & 0.452 & 0.330 & 0.205 & 0.149 & 0.452 \\
 & Kimi-K2.5 & \textbf{0.699} & \textbf{0.529} & \textbf{0.404} & 0.578 & \textbf{0.693} & \textbf{0.523} & \textbf{0.43} & \textbf{0.620} & \textbf{0.798} & \textbf{0.681} & \textbf{0.588} & \textbf{0.737} & \textbf{0.863} & \textbf{0.763} & \textbf{0.684} & \textbf{0.793} & \textbf{0.743} & \textbf{0.582} & \textbf{0.474} & 0.638 \\
\bottomrule
\end{tabular}%
}
\par\vspace{1mm}
\begin{minipage}{\textwidth}
\caption{L2 Domain (S4) results across Vietnamese (VI), Thai (TH), Indonesian (ID), Chinese (ZH), and Filipino (TL). Here, p@1 denotes pass@1, $p^2$ denotes pass$^2$, and $p^3$ denotes pass$^3$. Bold values indicate the best score within each domain--metric group. Qwen3-235B refers to \texttt{Qwen3-235B-A22B-Instruct-2507}.}
\end{minipage}
\end{table*}

\begin{table*}[!t]
\centering
\small
\begin{tabularx}{\textwidth}{@{}l >{\raggedright\arraybackslash}X@{}}
\toprule
\textbf{Level} & \textbf{Definition given to the judge} \\
\midrule
\multicolumn{2}{@{}l}{\textit{Agent errors}} \\
\texttt{critical} & The error directly caused task failure or violated important security/policy requirements, even if the task succeeded. \\
\texttt{minor} & The error was suboptimal but did not affect the final outcome or violate critical policies. \\
\midrule
\multicolumn{2}{@{}l}{\textit{Simulated-user errors}} \\
\texttt{critical\_helped} & The user error helped the agent succeed inappropriately, e.g.\ the user supplied information it should not have, making the task too easy. \\
\texttt{critical\_hindered} & The user error hindered the agent or made the task harder or impossible, e.g.\ the user supplied incorrect information or contradicted its instructions. \\
\texttt{minor} & The user made an error but it did not influence the simulation outcome. \\
\bottomrule
\end{tabularx}
\caption{Severity levels assigned by the LLM judge. We report \texttt{minor} as
\emph{Benign} throughout. Figures pool \texttt{critical\_helped} and
\texttt{critical\_hindered} into a single \emph{Critical} category for the simulated
user. Agent and user severities follow separate rubrics and are not directly
comparable: the user is additionally fact-checked on every factual detail it supplies,
with any detail not derivable from its instructions counted as a hallucination.}
\label{tab:severity-definitions}
\end{table*}
\begin{table*}[!t]
\centering
\small
\begin{tabularx}{\textwidth}{
    @{}
    l
    l
    >{\raggedright\arraybackslash}X
    @{}
}
\toprule
\textbf{Tag} & \textbf{Identifier} & \textbf{Definition given to the judge} \\
\midrule
\multicolumn{3}{@{}l}{\textit{Policy and procedure adherence}} \\
Missed Required Action & \texttt{missed\_required\_action} & Did not take a required action that was expected. \\
Guideline Violation & \texttt{guideline\_violation} & Message or action not consistent with the provided guidelines or policy. \\
Wrong Sequence & \texttt{wrong\_sequence} & Performed actions out of the expected order or sequence. \\
Premature Termination & \texttt{premature\_termination} & Ended the conversation or accepted an incomplete outcome while the other participant was actively working and making progress. Not applied when the user ended the conversation because the agent was unresponsive or repeatedly failing. \\
\midrule
\multicolumn{3}{@{}l}{\textit{Comprehension}} \\
Incorrect Interpretation & \texttt{incorrect\_interpretation} & Misinterpreted available information (e.g., misread a tool result or misunderstood a message). \\
Hallucination & \texttt{hallucination} & Made up information not grounded in guidelines, instructions, or tool call results. For user errors specifically: provided factual details (e.g., zip codes, sizes, product descriptions) not present in or derivable from the user instructions. \\
Inconsistent Behavior & \texttt{inconsistent\_behavior} & Action or statement contradicts earlier statements or actions in the conversation. \\
\midrule
\multicolumn{3}{@{}l}{\textit{Tool-schema logic}} \\
Tool Call Argument Error & \texttt{tool\_call\_argument\_error} & Made a tool call with correct schema but incorrect argument values. \\
Tool Call Schema Error & \texttt{tool\_call\_schema\_error} & Made a tool call with invalid tool name, missing arguments, or wrong argument types. \\
Irrelevant Tool Call & \texttt{irrelevant\_tool\_call} & Made a tool call not relevant to the current task or conversation state. \\
\midrule
\multicolumn{3}{@{}l}{\textit{Other}} \\
Revealed Info Early & \texttt{revealed\_info\_early} & Shared information before it was appropriate or before proper verification. \\
Other & \texttt{other} & Used only when no other tag applies; the judge describes the error type in its reasoning. \\
Interruption Error$^{\dagger}$ & \texttt{interruption\_error} & Interrupted inappropriately or failed to interrupt when appropriate. \\
\bottomrule
\end{tabularx}
\caption{Error tags available to the LLM judge, quoted from the review prompt. Groupings are ours and are used in
\Cref{sec:error-analysis} to relate tags to the capability a localized task stresses;
the judge receives a flat list. A single error instance may carry several tags, and may
span groups. $^{\dagger}$\texttt{interruption\_error} is defined only in the full-duplex
variant of the prompt; \seatau{} runs text-mode simulations exclusively, so this tag is
never offered to the judge and its count is zero by construction rather than by rarity.}
\label{tab:error-tag-definitions}
\end{table*}

\section{Detailed Error Analysis Results}
\label{sec:detailed-error}

\subsection{Error taxonomy}
\label{app:error-taxonomy}

Errors are annotated by an LLM judge following the review prompt of
\citet{shi2026tauknowledge}, as implemented in \taubench{}. For each error the judge
records the responsible party, one or more tags, a severity level, and the turn at which
it occurred. \Cref{tab:error-tag-definitions} lists the tags and
\Cref{tab:severity-definitions} the severity levels, both quoted from the prompt. Because
one error instance may carry several tags, tag counts sum to more than the number of
errors.

\subsection{Per-tag error rates}
\label{app:error-rates}

\begin{figure*}[!t]
  \centering
  \includegraphics[width=\textwidth]{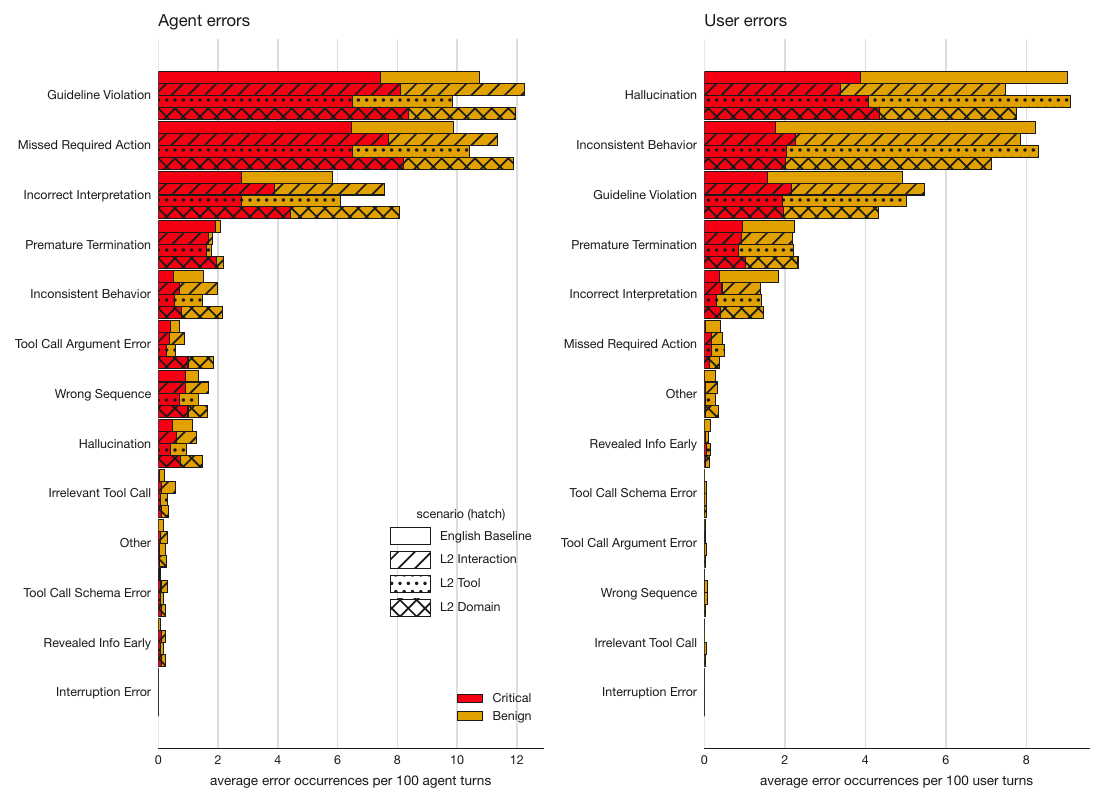}
  \caption{Agent and simulated-user error tags across the four scenarios.
    Bars give the average number of error occurrences per 100 turns of the relevant
    role --- agent errors per assistant turn, user errors per user turn --- computed
    within each scenario $\times$ language $\times$ domain cell and then averaged over
    the three domains and the five L2. Because a turn may be flagged more than once
    with the same tag, possibly at different severities, this counts occurrences
    rather than turns and sits slightly above the share of turns affected. For the simulated user,
    critical includes \texttt{critical\_helped} and \texttt{critical\_hindered}; agent and
    user severities follow role-specific rubrics and are not comparable across panels.
    All 13 tags are shown, with \texttt{interruption\_error} at zero in both panels
    because it applies only to full-duplex simulations. L2 Tool averages the five
    single-language tool variants; the mixed-language variants (Mix-2--Mix-5) appear in
    Appendix~\ref{app:error-counts}. Definitions are in
    \Cref{tab:error-tag-definitions,tab:severity-definitions}.
    Agent \texttt{gpt-5-mini}, simulated user
    \texttt{qwen3-235b-a22b-instruct-2507}, judge \texttt{DeepSeek-V4-Flash}.}
  \label{fig:error_tags_per_total_turns}
\end{figure*}
\subsection{Per-domain error counts}
\label{app:error-counts}

\Cref{tab:error-airline,tab:error-retail,tab:error-telecom} report simulation-level
severity per domain; the tables that follow give per-tag counts split by severity for the
agent and the simulated user. Both sets cover all four scenarios, the five
single-language tool variants, and the mixed-language tool variants (Mix-2--Mix-5).

% ============================================================
% Error Severity Tables
% ============================================================

% ------ AIRLINE (n=150 per condition) ------
\begin{table*}[!t]
\centering
\small
\begin{tabular}{ll ccc ccc}
\toprule
\multirow{2}{*}{\textbf{Setting}}
  & \multirow{2}{*}{\textbf{Variant}}
  & \multicolumn{3}{c}{\textbf{Agent Errors}}
  & \multicolumn{3}{c}{\textbf{User Errors}} \\
\cmidrule(lr){3-5}\cmidrule(lr){6-8}
  & & Critical & Minor & Correct & Critical & Minor & Correct \\
\midrule
  English Baseline & English & 81 & 22 & 47 & 38 & 48 & 64 \\
\midrule
\multirow{5}{*}{L2 Interaction}
   & Vietnamese & 93 & 25 & 32 & 43 & 48 & 59 \\
   & Thai & 105 & 14 & 31 & 42 & 45 & 63 \\
   & Indonesian & 88 & 18 & 44 & 36 & 49 & 65 \\
   & Chinese & 90 & 18 & 42 & 44 & 51 & 55 \\
   & Filipino & 94 & 18 & 38 & 53 & 36 & 61 \\
\midrule
\multirow{9}{*}{\makecell[l]{L2 Tool\\}}
   & VI Tool & 81 & 23 & 46 & 43 & 51 & 56 \\
   & TH Tool & 88 & 28 & 34 & 39 & 45 & 66 \\
   & ID Tool & 94 & 20 & 36 & 41 & 41 & 68 \\
   & ZH Tool & 84 & 23 & 43 & 35 & 44 & 71 \\
   & FIL Tools & 92 & 17 & 41 & 45 & 45 & 60 \\
   & Mix-2 (EN+TH) & 90 & 24 & 36 & 52 & 38 & 60 \\
   & Mix-3 (+VI) & 86 & 17 & 47 & 33 & 54 & 63 \\
   & Mix-4 (+ID) & 89 & 26 & 35 & 35 & 45 & 70 \\
   & Mix-5 (+ZH) & 102 & 15 & 33 & 40 & 41 & 69 \\
\midrule
\multirow{5}{*}{L2 Domain}
   & Vietnamese & 83 & 27 & 40 & 43 & 41 & 66 \\
   & Thai & 97 & 17 & 36 & 55 & 30 & 65 \\
   & Indonesian & 92 & 24 & 34 & 42 & 39 & 69 \\
   & Chinese & 103 & 15 & 32 & 40 & 48 & 62 \\
   & Filipino & 101 & 17 & 32 & 50 & 47 & 53 \\
\bottomrule
\end{tabular}
\caption{%
  Error analysis on the \textbf{Airline} domain (\texttt{gpt-5-mini} agent,
  \texttt{qwen3-235b} user simulator, \texttt{DeepSeek-V4-Flash} judge,
  $n{=}150$ per condition).
  Counts of simulations by maximum severity level.
}
\label{tab:error-airline}
\end{table*}

\begin{table*}[!t]
\centering
\small
\begin{tabular}{ll ccc ccc}
\toprule
\multirow{2}{*}{\textbf{Setting}}
  & \multirow{2}{*}{\textbf{Variant}}
  & \multicolumn{3}{c}{\textbf{Agent Errors}}
  & \multicolumn{3}{c}{\textbf{User Errors}} \\
\cmidrule(lr){3-5}\cmidrule(lr){6-8}
  & & Critical & Minor & Correct & Critical & Minor & Correct \\
\midrule
  English Baseline & English & 163 & 44 & 135 & 68 & 136 & 138 \\
\midrule
\multirow{5}{*}{L2 Interaction}
   & Vietnamese & 176 & 36 & 130 & 81 & 124 & 137 \\
   & Thai & 259 & 21 & 62 & 73 & 108 & 161 \\
   & Indonesian & 179 & 36 & 127 & 73 & 135 & 134 \\
   & Chinese & 161 & 43 & 138 & 70 & 142 & 130 \\
   & Filipino & 201 & 35 & 106 & 104 & 109 & 129 \\
\midrule
\multirow{9}{*}{\makecell[l]{L2\\Tool}}
   & VI Tools & 172 & 39 & 131 & 62 & 147 & 133 \\
   & TH Tools & 195 & 35 & 112 & 66 & 135 & 141 \\
   & ID Tools & 199 & 29 & 114 & 68 & 131 & 143 \\
   & ZH Tools & 171 & 44 & 127 & 76 & 129 & 137 \\
   & FIL Tools & 191 & 40 & 111 & 74 & 135 & 133 \\
   & Mix-2 (EN+TH) & 191 & 35 & 116 & 78 & 122 & 142 \\
   & Mix-3 (+VI) & 174 & 37 & 131 & 75 & 135 & 132 \\
   & Mix-4 (+ID) & 176 & 40 & 126 & 67 & 133 & 142 \\
   & Mix-5 (+ZH) & 197 & 32 & 113 & 62 & 140 & 140 \\
\midrule
\multirow{5}{*}{L2 Domain}
   & Vietnamese & 183 & 45 & 114 & 76 & 130 & 136 \\
   & Thai & 250 & 28 & 64 & 103 & 106 & 133 \\
   & Indonesian & 190 & 44 & 108 & 88 & 122 & 132 \\
   & Chinese & 175 & 41 & 126 & 78 & 135 & 129 \\
   & Filipino & 216 & 33 & 93 & 92 & 118 & 132 \\
\bottomrule
\end{tabular}
\caption{%
  Error analysis on the \textbf{Retail} domain (\texttt{gpt-5-mini} agent,
  \texttt{qwen3-235b} user simulator, \texttt{DeepSeek-V4-Flash} judge,
  $n{=}342$ per condition).
  Counts of simulations by maximum severity level.
}
\label{tab:error-retail}
\end{table*}
\vspace{-1em}
% ------ TELECOM (n=342 per condition) ------
\begin{table*}[!t]
\centering
\small
\begin{tabular}{ll ccc ccc}
\toprule
\multirow{2}{*}{\textbf{Setting}}
  & \multirow{2}{*}{\textbf{Variant}}
  & \multicolumn{3}{c}{\textbf{Agent Errors}}
  & \multicolumn{3}{c}{\textbf{User Errors}} \\
\cmidrule(lr){3-5}\cmidrule(lr){6-8}
  & & Critical & Minor & Correct & Critical & Minor & Correct \\
\midrule
  English Baseline & English & 123 & 36 & 183 & 78 & 116 & 148 \\
\midrule
\multirow{5}{*}{L2 Interaction}
   & Vietnamese & 151 & 43 & 148 & 90 & 111 & 141 \\
   & Thai & 158 & 40 & 144 & 88 & 118 & 136 \\
   & Indonesian & 158 & 40 & 144 & 78 & 116 & 148 \\
   & Chinese & 123 & 36 & 183 & 78 & 116 & 148 \\
   & Filipino & 147 & 32 & 163 & 107 & 87 & 148 \\
\midrule
\multirow{9}{*}{\makecell[l]{L2\\Tool}}
   & VI Tools & 132 & 40 & 170 & 75 & 124 & 143 \\
   & TH Tools & 135 & 39 & 168 & 77 & 124 & 141 \\
   & ID Tools & 157 & 31 & 154 & 77 & 122 & 143 \\
   & ZH Tools & 124 & 37 & 181 & 72 & 130 & 140 \\
   & FIL Tools & 143 & 31 & 168 & 85 & 103 & 154 \\
   & Mix-2 (EN+TH) & 146 & 31 & 165 & 82 & 117 & 143 \\
   & Mix-3 (+VI) & 147 & 35 & 160 & 81 & 120 & 141 \\
   & Mix-4 (+ID) & 145 & 34 & 163 & 84 & 119 & 139 \\
   & Mix-5 (+ZH) & 139 & 36 & 167 & 76 & 127 & 139 \\
\midrule
\multirow{5}{*}{L2 Domain}
   & Vietnamese & 166 & 40 & 136 & 86 & 115 & 141 \\
   & Thai & 170 & 36 & 136 & 84 & 109 & 149 \\
   & Indonesian & 195 & 38 & 109 & 87 & 107 & 148 \\
   & Chinese & 124 & 39 & 179 & 74 & 121 & 147 \\
   & Filipino & 140 & 44 & 158 & 96 & 106 & 140 \\
\bottomrule
\end{tabular}
\caption{%
  Error analysis on the \textbf{Telecom} domain (\texttt{gpt-5-mini} agent,
  \texttt{qwen3-235b} user simulator, \texttt{DeepSeek-V4-Flash} judge,
  $n{=}342$ per condition).
  Counts of simulations by maximum severity level.
}
\label{tab:error-telecom}
\end{table*}

% ============================================================
% Error Tag Tables -- split for readability
% Drop-in replacement for the original wide tables
% Requires booktabs (already required by the original tables).
% ============================================================

% ------ AIRLINE -- AGENT ERROR TAGS -- PART A ------
\begin{table*}[p]
\centering
\footnotesize
\renewcommand{\arraystretch}{1.05}
\setlength{\tabcolsep}{4pt}
\begin{tabular}{@{}lcccccc@{}}
\toprule
& \textbf{Eng.~Baseline} & \multicolumn{5}{c}{\textbf{L2 Interaction}} \\
\cmidrule(lr){2-2}\cmidrule(lr){3-7}
\textbf{Error Tag} & EN & VI & TH & ID & ZH & FIL \\
\midrule
  Guideline Violation & C:100 M:20 & C:116 M:38 & C:113 M:23 & C:89 M:31 & C:85 M:27 & C:110 M:28 \\
  Hallucination & C:6 M:7 & C:3 M:6 & C:4 M:8 & C:6 M:5 & C:9 M:1 & C:8 M:5 \\
  Inconsistent Behavior & C:4 M:8 & C:7 M:11 & C:11 M:9 & C:6 M:15 & C:5 M:7 & C:5 M:3 \\
  Incorrect Interpretation & C:33 M:25 & C:67 M:29 & C:56 M:30 & C:47 M:33 & C:46 M:38 & C:46 M:36 \\
  Interruption Error & C:0 M:0 & C:0 M:0 & C:0 M:0 & C:0 M:0 & C:0 M:0 & C:0 M:0 \\
  Irrelevant Tool Call & C:0 M:0 & C:3 M:4 & C:0 M:4 & C:0 M:4 & C:0 M:2 & C:0 M:0 \\
  Missed Required Action & C:71 M:16 & C:82 M:28 & C:98 M:18 & C:74 M:15 & C:85 M:20 & C:90 M:25 \\
  Other & C:0 M:1 & C:2 M:4 & C:0 M:0 & C:0 M:1 & C:1 M:3 & C:0 M:0 \\
  Premature Termination & C:28 M:1 & C:20 M:1 & C:30 M:2 & C:21 M:1 & C:18 M:0 & C:25 M:1 \\
  Revealed Info Early & C:0 M:1 & C:0 M:2 & C:0 M:1 & C:1 M:0 & C:1 M:0 & C:2 M:2 \\
  Tool Call Argument Error & C:5 M:4 & C:3 M:1 & C:7 M:6 & C:4 M:6 & C:3 M:7 & C:1 M:4 \\
  Tool Call Schema Error & C:0 M:0 & C:1 M:0 & C:2 M:2 & C:0 M:2 & C:0 M:1 & C:0 M:4 \\
  Wrong Sequence & C:9 M:3 & C:9 M:7 & C:8 M:3 & C:3 M:4 & C:11 M:4 & C:7 M:3 \\
\bottomrule
\end{tabular}
\caption{%
  Agent error tag counts --- \textbf{Airline} domain, English baseline and L2 Interaction settings
  (\texttt{gpt-5-mini} agent, \texttt{qwen3-235b} user simulator, \texttt{DeepSeek-V4-Flash} judge, $n{=}150$ per condition).
  \textbf{C:}~=~critical-severity count; \textbf{M:}~=~minor-severity count.
  Counts are judge-annotated error instances pooled over all simulations; a simulation may contribute several instances and an instance may carry several tags, so column sums can exceed $n$.
}
\label{tab:agent-tags-all-airline}
\end{table*}

% ------ AIRLINE -- AGENT ERROR TAGS -- PART B ------
\begin{table*}[p]
\centering
\footnotesize
\renewcommand{\arraystretch}{1.05}
\setlength{\tabcolsep}{4pt}
\begin{tabular}{@{}lccccc@{}}
\toprule
& \multicolumn{5}{c}{\textbf{L2 Tool --- Single-language}} \\
\cmidrule(lr){2-6}
\textbf{Error Tag} & VI & TH & ID & ZH & FIL \\
\midrule
  Guideline Violation & C:82 M:19 & C:105 M:30 & C:87 M:32 & C:92 M:26 & C:89 M:21 \\
  Hallucination & C:0 M:4 & C:5 M:2 & C:5 M:6 & C:2 M:4 & C:6 M:5 \\
  Inconsistent Behavior & C:12 M:5 & C:11 M:3 & C:5 M:6 & C:5 M:10 & C:3 M:5 \\
  Incorrect Interpretation & C:25 M:29 & C:46 M:35 & C:33 M:40 & C:42 M:31 & C:39 M:29 \\
  Interruption Error & C:0 M:0 & C:0 M:0 & C:0 M:0 & C:0 M:0 & C:0 M:0 \\
  Irrelevant Tool Call & C:4 M:0 & C:0 M:1 & C:0 M:0 & C:0 M:2 & C:0 M:1 \\
  Missed Required Action & C:79 M:22 & C:75 M:22 & C:78 M:17 & C:75 M:27 & C:85 M:22 \\
  Other & C:0 M:2 & C:0 M:1 & C:2 M:1 & C:0 M:0 & C:0 M:0 \\
  Premature Termination & C:22 M:1 & C:21 M:4 & C:19 M:1 & C:23 M:3 & C:34 M:1 \\
  Revealed Info Early & C:3 M:1 & C:1 M:1 & C:0 M:0 & C:0 M:0 & C:1 M:1 \\
  Tool Call Argument Error & C:2 M:6 & C:2 M:6 & C:1 M:4 & C:1 M:0 & C:5 M:0 \\
  Tool Call Schema Error & C:0 M:0 & C:0 M:1 & C:0 M:0 & C:2 M:0 & C:0 M:0 \\
  Wrong Sequence & C:12 M:6 & C:3 M:3 & C:5 M:3 & C:12 M:6 & C:5 M:2 \\
\bottomrule
\end{tabular}
\caption{%
  Agent error tag counts --- \textbf{Airline} domain, L2 Tool (single-language) settings.
}
\label{tab:agent-tags-all-airline-tool-l2}
\end{table*}

% ------ AIRLINE -- AGENT ERROR TAGS -- PART C ------
\begin{table*}[p]
\centering
\footnotesize
\renewcommand{\arraystretch}{1.05}
\setlength{\tabcolsep}{4pt}
\begin{tabular}{@{}lcccc@{}}
\toprule
& \multicolumn{4}{c}{\textbf{L2 Tool --- Mixed-language}} \\
\cmidrule(lr){2-5}
\textbf{Error Tag} & Mix-2 & Mix-3 & Mix-4 & Mix-5 \\
\midrule
  Guideline Violation & C:110 M:22 & C:97 M:16 & C:96 M:24 & C:103 M:26 \\
  Hallucination & C:3 M:2 & C:3 M:5 & C:2 M:5 & C:7 M:2 \\
  Inconsistent Behavior & C:3 M:5 & C:13 M:4 & C:8 M:9 & C:6 M:5 \\
  Incorrect Interpretation & C:34 M:29 & C:33 M:24 & C:36 M:34 & C:46 M:21 \\
  Interruption Error & C:0 M:0 & C:0 M:0 & C:0 M:0 & C:0 M:0 \\
  Irrelevant Tool Call & C:0 M:1 & C:1 M:1 & C:1 M:3 & C:0 M:2 \\
  Missed Required Action & C:78 M:20 & C:78 M:14 & C:87 M:33 & C:84 M:21 \\
  Other & C:0 M:1 & C:2 M:0 & C:1 M:1 & C:0 M:1 \\
  Premature Termination & C:29 M:1 & C:35 M:2 & C:26 M:1 & C:45 M:2 \\
  Revealed Info Early & C:1 M:1 & C:2 M:1 & C:0 M:0 & C:0 M:0 \\
  Tool Call Argument Error & C:3 M:3 & C:2 M:5 & C:1 M:4 & C:1 M:2 \\
  Tool Call Schema Error & C:0 M:1 & C:1 M:0 & C:0 M:1 & C:0 M:2 \\
  Wrong Sequence & C:5 M:1 & C:11 M:1 & C:5 M:1 & C:7 M:1 \\
\bottomrule
\end{tabular}
\caption{%
  Agent error tag counts --- \textbf{Airline} domain, L2 Tool (mixed-language) settings.
}
\label{tab:agent-tags-all-airline-tool-mix}
\end{table*}

% ------ AIRLINE -- AGENT ERROR TAGS -- PART D ------
\begin{table*}[p]
\centering
\footnotesize
\renewcommand{\arraystretch}{1.05}
\setlength{\tabcolsep}{4pt}
\begin{tabular}{@{}lccccc@{}}
\toprule
& \multicolumn{5}{c}{\textbf{L2 Domain}} \\
\cmidrule(lr){2-6}
\textbf{Error Tag} & VI & TH & ID & ZH & FIL \\
\midrule
  Guideline Violation & C:90 M:36 & C:130 M:28 & C:115 M:18 & C:108 M:21 & C:114 M:24 \\
  Hallucination & C:8 M:2 & C:14 M:10 & C:3 M:5 & C:7 M:8 & C:6 M:7 \\
  Inconsistent Behavior & C:6 M:13 & C:9 M:10 & C:6 M:8 & C:7 M:10 & C:6 M:5 \\
  Incorrect Interpretation & C:54 M:37 & C:71 M:35 & C:43 M:25 & C:47 M:32 & C:58 M:30 \\
  Interruption Error & C:0 M:0 & C:0 M:0 & C:0 M:0 & C:0 M:0 & C:0 M:0 \\
  Irrelevant Tool Call & C:2 M:0 & C:1 M:4 & C:1 M:0 & C:0 M:1 & C:0 M:1 \\
  Missed Required Action & C:77 M:15 & C:86 M:16 & C:96 M:25 & C:94 M:19 & C:110 M:23 \\
  Other & C:1 M:0 & C:1 M:0 & C:0 M:1 & C:0 M:1 & C:0 M:4 \\
  Premature Termination & C:21 M:9 & C:25 M:2 & C:28 M:0 & C:27 M:2 & C:30 M:1 \\
  Revealed Info Early & C:1 M:2 & C:0 M:2 & C:3 M:1 & C:1 M:2 & C:0 M:0 \\
  Tool Call Argument Error & C:5 M:3 & C:5 M:3 & C:7 M:8 & C:6 M:4 & C:6 M:2 \\
  Tool Call Schema Error & C:1 M:1 & C:0 M:0 & C:0 M:1 & C:0 M:1 & C:0 M:0 \\
  Wrong Sequence & C:8 M:3 & C:7 M:3 & C:13 M:5 & C:12 M:2 & C:11 M:5 \\
\bottomrule
\end{tabular}
\caption{%
  Agent error tag counts --- \textbf{Airline} domain, L2 Domain settings.
}
\label{tab:agent-tags-all-airline-domain}
\end{table*}

% ------ AIRLINE -- USER ERROR TAGS -- PART A ------
\begin{table*}[p]
\centering
\footnotesize
\renewcommand{\arraystretch}{1.05}
\setlength{\tabcolsep}{4pt}
\begin{tabular}{@{}lcccccc@{}}
\toprule
& \textbf{Eng.~Baseline} & \multicolumn{5}{c}{\textbf{L2 Interaction}} \\
\cmidrule(lr){2-2}\cmidrule(lr){3-7}
\textbf{Error Tag} & EN & VI & TH & ID & ZH & FIL \\
\midrule
  Guideline Violation & C:13 M:26 & C:18 M:22 & C:19 M:25 & C:15 M:30 & C:28 M:23 & C:26 M:31 \\
  Hallucination & C:26 M:27 & C:22 M:24 & C:31 M:24 & C:19 M:37 & C:23 M:31 & C:39 M:23 \\
  Inconsistent Behavior & C:15 M:34 & C:22 M:34 & C:15 M:28 & C:17 M:42 & C:25 M:35 & C:25 M:31 \\
  Incorrect Interpretation & C:4 M:7 & C:3 M:8 & C:2 M:9 & C:2 M:4 & C:3 M:9 & C:11 M:7 \\
  Interruption Error & C:0 M:0 & C:0 M:0 & C:0 M:0 & C:0 M:0 & C:0 M:0 & C:0 M:0 \\
  Irrelevant Tool Call & C:0 M:0 & C:0 M:0 & C:0 M:0 & C:0 M:0 & C:0 M:0 & C:0 M:0 \\
  Missed Required Action & C:0 M:1 & C:3 M:1 & C:2 M:0 & C:0 M:1 & C:1 M:2 & C:2 M:1 \\
  Other & C:0 M:1 & C:0 M:1 & C:0 M:0 & C:0 M:0 & C:0 M:2 & C:1 M:0 \\
  Premature Termination & C:9 M:10 & C:6 M:12 & C:12 M:15 & C:11 M:9 & C:11 M:18 & C:8 M:10 \\
  Revealed Info Early & C:0 M:2 & C:1 M:0 & C:0 M:2 & C:0 M:0 & C:0 M:1 & C:0 M:2 \\
  Tool Call Argument Error & C:0 M:0 & C:0 M:0 & C:0 M:0 & C:0 M:0 & C:0 M:0 & C:0 M:0 \\
  Tool Call Schema Error & C:0 M:0 & C:0 M:0 & C:0 M:0 & C:0 M:0 & C:0 M:0 & C:0 M:0 \\
  Wrong Sequence & C:0 M:0 & C:0 M:0 & C:0 M:0 & C:0 M:1 & C:0 M:0 & C:0 M:0 \\
\bottomrule
\end{tabular}
\caption{%
  User error tag counts --- \textbf{Airline} domain, English baseline and L2 Interaction settings
  (\texttt{gpt-5-mini} agent, \texttt{qwen3-235b} user simulator, \texttt{DeepSeek-V4-Flash} judge, $n{=}150$ per condition).
  \textbf{C:}~=~critical-severity count; \textbf{M:}~=~minor-severity count.
  For the simulated user, \textbf{C:} pools \texttt{critical\_helped} and \texttt{critical\_hindered}.
  Counts are judge-annotated error instances pooled over all simulations; a simulation may contribute several instances and an instance may carry several tags, so column sums can exceed $n$.
}
\label{tab:user-tags-all-airline}
\end{table*}

% ------ AIRLINE -- USER ERROR TAGS -- PART B ------
\begin{table*}[p]
\centering
\footnotesize
\renewcommand{\arraystretch}{1.05}
\setlength{\tabcolsep}{4pt}
\begin{tabular}{@{}lccccc@{}}
\toprule
& \multicolumn{5}{c}{\textbf{L2 Tool --- Single-language}} \\
\cmidrule(lr){2-6}
\textbf{Error Tag} & VI & TH & ID & ZH & FIL \\
\midrule
  Guideline Violation & C:19 M:36 & C:9 M:17 & C:16 M:13 & C:14 M:25 & C:22 M:18 \\
  Hallucination & C:39 M:28 & C:39 M:20 & C:25 M:21 & C:32 M:28 & C:35 M:29 \\
  Inconsistent Behavior & C:15 M:45 & C:19 M:27 & C:13 M:40 & C:14 M:37 & C:24 M:33 \\
  Incorrect Interpretation & C:2 M:11 & C:3 M:2 & C:3 M:5 & C:1 M:6 & C:1 M:4 \\
  Interruption Error & C:0 M:0 & C:0 M:0 & C:0 M:0 & C:0 M:0 & C:0 M:0 \\
  Irrelevant Tool Call & C:0 M:0 & C:0 M:0 & C:0 M:0 & C:0 M:0 & C:0 M:0 \\
  Missed Required Action & C:1 M:1 & C:1 M:0 & C:3 M:1 & C:1 M:0 & C:1 M:2 \\
  Other & C:0 M:0 & C:0 M:2 & C:0 M:0 & C:1 M:0 & C:0 M:1 \\
  Premature Termination & C:4 M:7 & C:10 M:17 & C:10 M:16 & C:7 M:10 & C:9 M:15 \\
  Revealed Info Early & C:0 M:2 & C:0 M:0 & C:0 M:0 & C:1 M:1 & C:0 M:1 \\
  Tool Call Argument Error & C:0 M:0 & C:0 M:0 & C:0 M:0 & C:0 M:0 & C:0 M:0 \\
  Tool Call Schema Error & C:0 M:0 & C:0 M:0 & C:0 M:0 & C:0 M:0 & C:0 M:0 \\
  Wrong Sequence & C:0 M:0 & C:0 M:0 & C:0 M:0 & C:0 M:1 & C:0 M:1 \\
\bottomrule
\end{tabular}
\caption{%
  User error tag counts --- \textbf{Airline} domain, L2 Tool (single-language) settings.
}
\label{tab:user-tags-all-airline-tool-l2}
\end{table*}

% ------ AIRLINE -- USER ERROR TAGS -- PART C ------
\begin{table*}[p]
\centering
\footnotesize
\renewcommand{\arraystretch}{1.05}
\setlength{\tabcolsep}{4pt}
\begin{tabular}{@{}lcccc@{}}
\toprule
& \multicolumn{4}{c}{\textbf{L2 Tool --- Mixed-language}} \\
\cmidrule(lr){2-5}
\textbf{Error Tag} & Mix-2 & Mix-3 & Mix-4 & Mix-5 \\
\midrule
  Guideline Violation & C:23 M:19 & C:16 M:22 & C:19 M:20 & C:18 M:14 \\
  Hallucination & C:46 M:19 & C:30 M:26 & C:25 M:21 & C:33 M:20 \\
  Inconsistent Behavior & C:21 M:37 & C:19 M:36 & C:15 M:37 & C:25 M:33 \\
  Incorrect Interpretation & C:6 M:8 & C:4 M:9 & C:3 M:4 & C:1 M:7 \\
  Interruption Error & C:0 M:0 & C:0 M:0 & C:0 M:0 & C:0 M:0 \\
  Irrelevant Tool Call & C:0 M:0 & C:0 M:0 & C:0 M:0 & C:0 M:0 \\
  Missed Required Action & C:1 M:1 & C:0 M:0 & C:1 M:3 & C:0 M:1 \\
  Other & C:0 M:0 & C:0 M:3 & C:1 M:0 & C:0 M:0 \\
  Premature Termination & C:9 M:7 & C:7 M:9 & C:6 M:11 & C:3 M:11 \\
  Revealed Info Early & C:0 M:1 & C:0 M:0 & C:0 M:0 & C:0 M:1 \\
  Tool Call Argument Error & C:0 M:0 & C:0 M:0 & C:0 M:0 & C:0 M:0 \\
  Tool Call Schema Error & C:0 M:0 & C:0 M:0 & C:0 M:0 & C:0 M:0 \\
  Wrong Sequence & C:0 M:0 & C:1 M:0 & C:0 M:0 & C:0 M:0 \\
\bottomrule
\end{tabular}
\caption{%
  User error tag counts --- \textbf{Airline} domain, L2 Tool (mixed-language) settings.
}
\label{tab:user-tags-all-airline-tool-mix}
\end{table*}

% ------ AIRLINE -- USER ERROR TAGS -- PART D ------
\begin{table*}[p]
\centering
\footnotesize
\renewcommand{\arraystretch}{1.05}
\setlength{\tabcolsep}{4pt}
\begin{tabular}{@{}lccccc@{}}
\toprule
& \multicolumn{5}{c}{\textbf{L2 Domain}} \\
\cmidrule(lr){2-6}
\textbf{Error Tag} & VI & TH & ID & ZH & FIL \\
\midrule
  Guideline Violation & C:16 M:20 & C:29 M:14 & C:20 M:13 & C:18 M:19 & C:21 M:24 \\
  Hallucination & C:35 M:22 & C:48 M:13 & C:24 M:13 & C:33 M:18 & C:29 M:21 \\
  Inconsistent Behavior & C:16 M:23 & C:26 M:30 & C:15 M:27 & C:18 M:37 & C:19 M:39 \\
  Incorrect Interpretation & C:3 M:8 & C:3 M:2 & C:3 M:6 & C:0 M:13 & C:5 M:8 \\
  Interruption Error & C:0 M:0 & C:0 M:0 & C:0 M:0 & C:0 M:0 & C:0 M:0 \\
  Irrelevant Tool Call & C:0 M:0 & C:0 M:0 & C:0 M:0 & C:0 M:0 & C:0 M:0 \\
  Missed Required Action & C:2 M:3 & C:0 M:1 & C:0 M:0 & C:0 M:0 & C:3 M:3 \\
  Other & C:0 M:2 & C:0 M:0 & C:0 M:1 & C:0 M:0 & C:2 M:1 \\
  Premature Termination & C:8 M:14 & C:10 M:8 & C:9 M:11 & C:9 M:12 & C:11 M:16 \\
  Revealed Info Early & C:0 M:3 & C:0 M:0 & C:0 M:0 & C:0 M:3 & C:0 M:1 \\
  Tool Call Argument Error & C:0 M:0 & C:0 M:0 & C:0 M:0 & C:0 M:0 & C:0 M:0 \\
  Tool Call Schema Error & C:0 M:0 & C:0 M:0 & C:0 M:0 & C:0 M:0 & C:0 M:0 \\
  Wrong Sequence & C:0 M:0 & C:0 M:0 & C:0 M:0 & C:0 M:1 & C:0 M:0 \\
\bottomrule
\end{tabular}
\caption{%
  User error tag counts --- \textbf{Airline} domain, L2 Domain settings.
}
\label{tab:user-tags-all-airline-domain}
\end{table*}

% ------ RETAIL -- AGENT ERROR TAGS -- PART A ------
\begin{table*}[p]
\centering
\footnotesize
\renewcommand{\arraystretch}{1.05}
\setlength{\tabcolsep}{4pt}
\begin{tabular}{@{}lcccccc@{}}
\toprule
& \textbf{Eng.~Baseline} & \multicolumn{5}{c}{\textbf{L2 Interaction}} \\
\cmidrule(lr){2-2}\cmidrule(lr){3-7}
\textbf{Error Tag} & EN & VI & TH & ID & ZH & FIL \\
\midrule
  Guideline Violation & C:70 M:74 & C:99 M:80 & C:105 M:81 & C:88 M:83 & C:71 M:59 & C:86 M:94 \\
  Hallucination & C:7 M:11 & C:9 M:12 & C:15 M:13 & C:17 M:17 & C:13 M:11 & C:10 M:10 \\
  Inconsistent Behavior & C:10 M:15 & C:17 M:24 & C:13 M:24 & C:8 M:15 & C:8 M:28 & C:8 M:16 \\
  Incorrect Interpretation & C:34 M:39 & C:32 M:54 & C:42 M:59 & C:38 M:54 & C:26 M:44 & C:37 M:39 \\
  Interruption Error & C:0 M:0 & C:0 M:0 & C:0 M:0 & C:0 M:0 & C:0 M:0 & C:0 M:0 \\
  Irrelevant Tool Call & C:2 M:3 & C:1 M:3 & C:1 M:4 & C:5 M:5 & C:0 M:5 & C:0 M:2 \\
  Missed Required Action & C:75 M:53 & C:78 M:62 & C:95 M:51 & C:88 M:55 & C:93 M:53 & C:94 M:63 \\
  Other & C:0 M:2 & C:0 M:3 & C:1 M:7 & C:2 M:1 & C:0 M:3 & C:0 M:3 \\
  Premature Termination & C:10 M:5 & C:5 M:2 & C:7 M:2 & C:11 M:0 & C:5 M:7 & C:9 M:5 \\
  Revealed Info Early & C:0 M:0 & C:1 M:5 & C:4 M:4 & C:4 M:2 & C:1 M:2 & C:2 M:3 \\
  Tool Call Argument Error & C:8 M:5 & C:7 M:9 & C:7 M:18 & C:8 M:6 & C:9 M:10 & C:5 M:3 \\
  Tool Call Schema Error & C:1 M:2 & C:3 M:5 & C:4 M:6 & C:1 M:5 & C:0 M:4 & C:4 M:9 \\
  Wrong Sequence & C:20 M:7 & C:24 M:24 & C:20 M:17 & C:19 M:20 & C:19 M:16 & C:23 M:15 \\
\bottomrule
\end{tabular}
\caption{%
  Agent error tag counts --- \textbf{Retail} domain, English baseline and L2 Interaction settings
  (\texttt{gpt-5-mini} agent, \texttt{qwen3-235b} user simulator, \texttt{DeepSeek-V4-Flash} judge, $n{=}342$ per condition).
  \textbf{C:}~=~critical-severity count; \textbf{M:}~=~minor-severity count.
  Counts are judge-annotated error instances pooled over all simulations; a simulation may contribute several instances and an instance may carry several tags, so column sums can exceed $n$.
}
\label{tab:agent-tags-all-retail}
\end{table*}

% ------ RETAIL -- AGENT ERROR TAGS -- PART B ------
\begin{table*}[p]
\centering
\footnotesize
\renewcommand{\arraystretch}{1.05}
\setlength{\tabcolsep}{4pt}
\begin{tabular}{@{}lccccc@{}}
\toprule
& \multicolumn{5}{c}{\textbf{L2 Tool --- Single-language}} \\
\cmidrule(lr){2-6}
\textbf{Error Tag} & VI & TH & ID & ZH & FIL \\
\midrule
  Guideline Violation & C:68 M:52 & C:70 M:66 & C:69 M:69 & C:62 M:55 & C:78 M:56 \\
  Hallucination & C:9 M:11 & C:6 M:13 & C:3 M:11 & C:16 M:7 & C:15 M:10 \\
  Inconsistent Behavior & C:10 M:15 & C:6 M:14 & C:7 M:17 & C:8 M:10 & C:5 M:21 \\
  Incorrect Interpretation & C:31 M:53 & C:40 M:47 & C:28 M:39 & C:30 M:40 & C:36 M:34 \\
  Interruption Error & C:0 M:0 & C:0 M:0 & C:0 M:0 & C:0 M:0 & C:0 M:0 \\
  Irrelevant Tool Call & C:1 M:3 & C:1 M:5 & C:0 M:0 & C:0 M:2 & C:2 M:2 \\
  Missed Required Action & C:62 M:59 & C:79 M:66 & C:74 M:64 & C:69 M:63 & C:69 M:69 \\
  Other & C:0 M:3 & C:0 M:3 & C:0 M:4 & C:2 M:1 & C:1 M:2 \\
  Premature Termination & C:6 M:3 & C:7 M:3 & C:8 M:3 & C:12 M:3 & C:6 M:1 \\
  Revealed Info Early & C:2 M:3 & C:1 M:2 & C:1 M:5 & C:0 M:6 & C:0 M:0 \\
  Tool Call Argument Error & C:9 M:7 & C:5 M:7 & C:8 M:9 & C:6 M:3 & C:8 M:6 \\
  Tool Call Schema Error & C:0 M:6 & C:2 M:6 & C:1 M:8 & C:4 M:2 & C:4 M:5 \\
  Wrong Sequence & C:16 M:14 & C:12 M:21 & C:18 M:18 & C:8 M:13 & C:11 M:9 \\
\bottomrule
\end{tabular}
\caption{%
  Agent error tag counts --- \textbf{Retail} domain, L2 Tool (single-language) settings.
}
\label{tab:agent-tags-all-retail-tool-l2}
\end{table*}

% ------ RETAIL -- AGENT ERROR TAGS -- PART C ------
\begin{table*}[p]
\centering
\footnotesize
\renewcommand{\arraystretch}{1.05}
\setlength{\tabcolsep}{4pt}
\begin{tabular}{@{}lcccc@{}}
\toprule
& \multicolumn{4}{c}{\textbf{L2 Tool --- Mixed-language}} \\
\cmidrule(lr){2-5}
\textbf{Error Tag} & Mix-2 & Mix-3 & Mix-4 & Mix-5 \\
\midrule
  Guideline Violation & C:49 M:55 & C:67 M:48 & C:63 M:53 & C:78 M:72 \\
  Hallucination & C:11 M:10 & C:11 M:12 & C:17 M:18 & C:12 M:18 \\
  Inconsistent Behavior & C:10 M:15 & C:10 M:13 & C:7 M:16 & C:8 M:26 \\
  Incorrect Interpretation & C:22 M:33 & C:29 M:32 & C:38 M:43 & C:27 M:65 \\
  Interruption Error & C:0 M:0 & C:0 M:0 & C:0 M:0 & C:0 M:0 \\
  Irrelevant Tool Call & C:0 M:1 & C:1 M:3 & C:0 M:3 & C:1 M:5 \\
  Missed Required Action & C:58 M:56 & C:63 M:71 & C:73 M:49 & C:59 M:67 \\
  Other & C:1 M:2 & C:0 M:4 & C:0 M:1 & C:0 M:6 \\
  Premature Termination & C:12 M:4 & C:17 M:2 & C:5 M:1 & C:6 M:2 \\
  Revealed Info Early & C:0 M:4 & C:3 M:1 & C:0 M:5 & C:1 M:0 \\
  Tool Call Argument Error & C:5 M:14 & C:7 M:4 & C:8 M:9 & C:5 M:8 \\
  Tool Call Schema Error & C:1 M:2 & C:0 M:9 & C:4 M:5 & C:2 M:7 \\
  Wrong Sequence & C:14 M:14 & C:7 M:10 & C:11 M:13 & C:11 M:18 \\
\bottomrule
\end{tabular}
\caption{%
  Agent error tag counts --- \textbf{Retail} domain, L2 Tool (mixed-language) settings.
}
\label{tab:agent-tags-all-retail-tool-mix}
\end{table*}

% ------ RETAIL -- AGENT ERROR TAGS -- PART D ------
\begin{table*}[p]
\centering
\footnotesize
\renewcommand{\arraystretch}{1.05}
\setlength{\tabcolsep}{4pt}
\begin{tabular}{@{}lccccc@{}}
\toprule
& \multicolumn{5}{c}{\textbf{L2 Domain}} \\
\cmidrule(lr){2-6}
\textbf{Error Tag} & VI & TH & ID & ZH & FIL \\
\midrule
  Guideline Violation & C:73 M:76 & C:128 M:54 & C:92 M:53 & C:112 M:93 & C:92 M:59 \\
  Hallucination & C:18 M:17 & C:30 M:10 & C:18 M:17 & C:9 M:18 & C:14 M:13 \\
  Inconsistent Behavior & C:13 M:27 & C:29 M:25 & C:12 M:22 & C:24 M:27 & C:22 M:22 \\
  Incorrect Interpretation & C:48 M:88 & C:142 M:62 & C:50 M:56 & C:63 M:63 & C:50 M:55 \\
  Interruption Error & C:0 M:0 & C:0 M:0 & C:0 M:0 & C:0 M:0 & C:0 M:0 \\
  Irrelevant Tool Call & C:0 M:1 & C:11 M:1 & C:1 M:4 & C:1 M:5 & C:1 M:1 \\
  Missed Required Action & C:73 M:66 & C:160 M:62 & C:75 M:65 & C:98 M:83 & C:101 M:81 \\
  Other & C:0 M:3 & C:1 M:5 & C:1 M:2 & C:1 M:5 & C:2 M:1 \\
  Premature Termination & C:12 M:4 & C:28 M:4 & C:15 M:4 & C:14 M:1 & C:7 M:3 \\
  Revealed Info Early & C:1 M:3 & C:2 M:3 & C:2 M:3 & C:0 M:3 & C:2 M:2 \\
  Tool Call Argument Error & C:35 M:49 & C:48 M:39 & C:39 M:28 & C:48 M:25 & C:37 M:34 \\
  Tool Call Schema Error & C:6 M:1 & C:3 M:9 & C:5 M:9 & C:6 M:5 & C:2 M:4 \\
  Wrong Sequence & C:14 M:11 & C:24 M:8 & C:19 M:21 & C:12 M:15 & C:13 M:12 \\
\bottomrule
\end{tabular}
\caption{%
  Agent error tag counts --- \textbf{Retail} domain, L2 Domain settings.
}
\label{tab:agent-tags-all-retail-domain}
\end{table*}

% ------ RETAIL -- USER ERROR TAGS -- PART A ------
\begin{table*}[p]
\centering
\footnotesize
\renewcommand{\arraystretch}{1.05}
\setlength{\tabcolsep}{4pt}
\begin{tabular}{@{}lcccccc@{}}
\toprule
& \textbf{Eng.~Baseline} & \multicolumn{5}{c}{\textbf{L2 Interaction}} \\
\cmidrule(lr){2-2}\cmidrule(lr){3-7}
\textbf{Error Tag} & EN & VI & TH & ID & ZH & FIL \\
\midrule
  Guideline Violation & C:19 M:50 & C:32 M:51 & C:41 M:72 & C:28 M:36 & C:38 M:42 & C:31 M:56 \\
  Hallucination & C:69 M:105 & C:87 M:91 & C:71 M:82 & C:59 M:92 & C:55 M:86 & C:95 M:108 \\
  Inconsistent Behavior & C:22 M:93 & C:38 M:85 & C:40 M:101 & C:38 M:96 & C:32 M:81 & C:44 M:101 \\
  Incorrect Interpretation & C:4 M:21 & C:15 M:8 & C:9 M:10 & C:6 M:8 & C:1 M:13 & C:10 M:7 \\
  Interruption Error & C:0 M:0 & C:0 M:0 & C:0 M:0 & C:0 M:0 & C:0 M:0 & C:0 M:0 \\
  Irrelevant Tool Call & C:0 M:0 & C:0 M:0 & C:0 M:0 & C:0 M:0 & C:0 M:0 & C:0 M:0 \\
  Missed Required Action & C:0 M:2 & C:1 M:1 & C:2 M:4 & C:4 M:2 & C:3 M:1 & C:3 M:3 \\
  Other & C:0 M:4 & C:2 M:3 & C:2 M:10 & C:1 M:1 & C:0 M:8 & C:1 M:1 \\
  Premature Termination & C:9 M:19 & C:11 M:17 & C:17 M:18 & C:12 M:19 & C:16 M:22 & C:10 M:15 \\
  Revealed Info Early & C:0 M:1 & C:0 M:0 & C:2 M:0 & C:0 M:0 & C:0 M:0 & C:0 M:0 \\
  Tool Call Argument Error & C:0 M:0 & C:0 M:0 & C:0 M:0 & C:0 M:0 & C:0 M:0 & C:0 M:0 \\
  Tool Call Schema Error & C:0 M:0 & C:0 M:0 & C:0 M:0 & C:0 M:0 & C:0 M:0 & C:0 M:0 \\
  Wrong Sequence & C:0 M:0 & C:0 M:0 & C:0 M:0 & C:0 M:0 & C:0 M:0 & C:0 M:0 \\
\bottomrule
\end{tabular}
\caption{%
  User error tag counts --- \textbf{Retail} domain, English baseline and L2 Interaction settings
  (\texttt{gpt-5-mini} agent, \texttt{qwen3-235b} user simulator, \texttt{DeepSeek-V4-Flash} judge, $n{=}342$ per condition).
  \textbf{C:}~=~critical-severity count; \textbf{M:}~=~minor-severity count.
  For the simulated user, \textbf{C:} pools \texttt{critical\_helped} and \texttt{critical\_hindered}.
  Counts are judge-annotated error instances pooled over all simulations; a simulation may contribute several instances and an instance may carry several tags, so column sums can exceed $n$.
}
\label{tab:user-tags-all-retail}
\end{table*}

% ------ RETAIL -- USER ERROR TAGS -- PART B ------
\begin{table*}[p]
\centering
\footnotesize
\renewcommand{\arraystretch}{1.05}
\setlength{\tabcolsep}{4pt}
\begin{tabular}{@{}lccccc@{}}
\toprule
& \multicolumn{5}{c}{\textbf{L2 Tool --- Single-language}} \\
\cmidrule(lr){2-6}
\textbf{Error Tag} & VI & TH & ID & ZH & FIL \\
\midrule
  Guideline Violation & C:19 M:46 & C:22 M:34 & C:41 M:43 & C:29 M:57 & C:24 M:47 \\
  Hallucination & C:71 M:115 & C:55 M:90 & C:61 M:132 & C:74 M:104 & C:54 M:123 \\
  Inconsistent Behavior & C:28 M:99 & C:28 M:86 & C:27 M:86 & C:30 M:103 & C:37 M:92 \\
  Incorrect Interpretation & C:2 M:10 & C:5 M:16 & C:4 M:8 & C:8 M:16 & C:7 M:15 \\
  Interruption Error & C:0 M:0 & C:0 M:0 & C:0 M:0 & C:0 M:0 & C:0 M:0 \\
  Irrelevant Tool Call & C:0 M:0 & C:0 M:0 & C:0 M:0 & C:0 M:0 & C:0 M:0 \\
  Missed Required Action & C:2 M:1 & C:5 M:4 & C:0 M:4 & C:3 M:4 & C:2 M:3 \\
  Other & C:0 M:2 & C:0 M:4 & C:0 M:2 & C:0 M:2 & C:1 M:3 \\
  Premature Termination & C:10 M:16 & C:7 M:15 & C:15 M:14 & C:10 M:16 & C:14 M:17 \\
  Revealed Info Early & C:2 M:2 & C:1 M:0 & C:3 M:0 & C:1 M:2 & C:0 M:1 \\
  Tool Call Argument Error & C:0 M:0 & C:0 M:0 & C:0 M:0 & C:0 M:0 & C:0 M:0 \\
  Tool Call Schema Error & C:0 M:0 & C:0 M:0 & C:0 M:0 & C:0 M:0 & C:0 M:0 \\
  Wrong Sequence & C:0 M:0 & C:0 M:0 & C:0 M:0 & C:0 M:0 & C:0 M:0 \\
\bottomrule
\end{tabular}
\caption{%
  User error tag counts --- \textbf{Retail} domain, L2 Tool (single-language) settings.
}
\label{tab:user-tags-all-retail-tool-l2}
\end{table*}

% ------ RETAIL -- USER ERROR TAGS -- PART C ------
\begin{table*}[p]
\centering
\footnotesize
\renewcommand{\arraystretch}{1.05}
\setlength{\tabcolsep}{4pt}
\begin{tabular}{@{}lcccc@{}}
\toprule
& \multicolumn{4}{c}{\textbf{L2 Tool --- Mixed-language}} \\
\cmidrule(lr){2-5}
\textbf{Error Tag} & Mix-2 & Mix-3 & Mix-4 & Mix-5 \\
\midrule
  Guideline Violation & C:17 M:37 & C:30 M:39 & C:23 M:29 & C:27 M:40 \\
  Hallucination & C:53 M:123 & C:82 M:106 & C:69 M:94 & C:64 M:129 \\
  Inconsistent Behavior & C:14 M:66 & C:38 M:94 & C:31 M:76 & C:21 M:85 \\
  Incorrect Interpretation & C:7 M:12 & C:5 M:12 & C:5 M:12 & C:12 M:14 \\
  Interruption Error & C:0 M:0 & C:0 M:0 & C:0 M:0 & C:0 M:0 \\
  Irrelevant Tool Call & C:0 M:0 & C:0 M:0 & C:0 M:0 & C:0 M:0 \\
  Missed Required Action & C:0 M:0 & C:0 M:6 & C:1 M:2 & C:2 M:2 \\
  Other & C:1 M:5 & C:0 M:7 & C:0 M:2 & C:0 M:6 \\
  Premature Termination & C:6 M:14 & C:10 M:15 & C:17 M:12 & C:11 M:21 \\
  Revealed Info Early & C:1 M:1 & C:1 M:1 & C:0 M:0 & C:0 M:0 \\
  Tool Call Argument Error & C:0 M:0 & C:0 M:0 & C:0 M:0 & C:0 M:0 \\
  Tool Call Schema Error & C:0 M:0 & C:0 M:0 & C:0 M:0 & C:0 M:0 \\
  Wrong Sequence & C:0 M:1 & C:0 M:0 & C:0 M:0 & C:0 M:1 \\
\bottomrule
\end{tabular}
\caption{%
  User error tag counts --- \textbf{Retail} domain, L2 Tool (mixed-language) settings.
}
\label{tab:user-tags-all-retail-tool-mix}
\end{table*}

% ------ RETAIL -- USER ERROR TAGS -- PART D ------
\begin{table*}[p]
\centering
\footnotesize
\renewcommand{\arraystretch}{1.05}
\setlength{\tabcolsep}{4pt}
\begin{tabular}{@{}lccccc@{}}
\toprule
& \multicolumn{5}{c}{\textbf{L2 Domain}} \\
\cmidrule(lr){2-6}
\textbf{Error Tag} & VI & TH & ID & ZH & FIL \\
\midrule
  Guideline Violation & C:25 M:31 & C:29 M:23 & C:29 M:32 & C:33 M:31 & C:23 M:33 \\
  Hallucination & C:75 M:76 & C:111 M:74 & C:76 M:105 & C:119 M:110 & C:115 M:106 \\
  Inconsistent Behavior & C:39 M:94 & C:39 M:86 & C:34 M:102 & C:26 M:84 & C:51 M:111 \\
  Incorrect Interpretation & C:5 M:20 & C:11 M:16 & C:7 M:12 & C:5 M:17 & C:13 M:15 \\
  Interruption Error & C:0 M:0 & C:0 M:0 & C:0 M:0 & C:0 M:0 & C:0 M:0 \\
  Irrelevant Tool Call & C:0 M:0 & C:0 M:0 & C:0 M:0 & C:0 M:0 & C:0 M:0 \\
  Missed Required Action & C:2 M:0 & C:3 M:2 & C:2 M:5 & C:2 M:3 & C:3 M:2 \\
  Other & C:0 M:4 & C:0 M:3 & C:0 M:4 & C:1 M:3 & C:0 M:4 \\
  Premature Termination & C:8 M:18 & C:28 M:24 & C:16 M:19 & C:11 M:12 & C:22 M:16 \\
  Revealed Info Early & C:0 M:1 & C:3 M:2 & C:0 M:0 & C:0 M:1 & C:0 M:2 \\
  Tool Call Argument Error & C:0 M:0 & C:0 M:0 & C:0 M:0 & C:0 M:0 & C:0 M:0 \\
  Tool Call Schema Error & C:0 M:0 & C:0 M:0 & C:0 M:0 & C:0 M:0 & C:0 M:0 \\
  Wrong Sequence & C:0 M:0 & C:0 M:0 & C:0 M:0 & C:0 M:0 & C:0 M:0 \\
\bottomrule
\end{tabular}
\caption{%
  User error tag counts --- \textbf{Retail} domain, L2 Domain settings.
}
\label{tab:user-tags-all-retail-domain}
\end{table*}

% ------ TELECOM -- AGENT ERROR TAGS -- PART A ------
\begin{table*}[p]
\centering
\footnotesize
\renewcommand{\arraystretch}{1.05}
\setlength{\tabcolsep}{4pt}
\begin{tabular}{@{}lcccccc@{}}
\toprule
& \textbf{Eng.~Baseline} & \multicolumn{5}{c}{\textbf{L2 Interaction}} \\
\cmidrule(lr){2-2}\cmidrule(lr){3-7}
\textbf{Error Tag} & EN & VI & TH & ID & ZH & FIL \\
\midrule
  Guideline Violation & C:88 M:81 & C:77 M:128 & C:99 M:105 & C:103 M:116 & C:71 M:46 & C:61 M:68 \\
  Hallucination & C:4 M:8 & C:6 M:16 & C:7 M:15 & C:6 M:14 & C:3 M:5 & C:4 M:11 \\
  Inconsistent Behavior & C:9 M:28 & C:15 M:49 & C:16 M:29 & C:12 M:40 & C:4 M:20 & C:6 M:32 \\
  Incorrect Interpretation & C:39 M:94 & C:38 M:84 & C:47 M:100 & C:43 M:100 & C:36 M:53 & C:29 M:60 \\
  Interruption Error & C:0 M:0 & C:0 M:0 & C:0 M:0 & C:0 M:0 & C:0 M:0 & C:0 M:0 \\
  Irrelevant Tool Call & C:0 M:10 & C:4 M:36 & C:7 M:24 & C:5 M:19 & C:1 M:6 & C:3 M:18 \\
  Missed Required Action & C:127 M:144 & C:106 M:132 & C:142 M:118 & C:138 M:132 & C:109 M:107 & C:107 M:97 \\
  Other & C:0 M:6 & C:1 M:6 & C:0 M:10 & C:1 M:8 & C:1 M:3 & C:1 M:8 \\
  Premature Termination & C:25 M:4 & C:25 M:6 & C:36 M:4 & C:28 M:2 & C:23 M:2 & C:18 M:1 \\
  Revealed Info Early & C:0 M:2 & C:2 M:2 & C:2 M:2 & C:0 M:1 & C:1 M:1 & C:0 M:0 \\
  Tool Call Argument Error & C:1 M:0 & C:6 M:7 & C:4 M:19 & C:0 M:4 & C:1 M:3 & C:1 M:4 \\
  Tool Call Schema Error & C:0 M:1 & C:1 M:3 & C:0 M:6 & C:0 M:1 & C:1 M:0 & C:1 M:2 \\
  Wrong Sequence & C:6 M:12 & C:13 M:24 & C:7 M:27 & C:10 M:16 & C:12 M:7 & C:5 M:8 \\
\bottomrule
\end{tabular}
\caption{%
  Agent error tag counts --- \textbf{Telecom} domain, English baseline and L2 Interaction settings
  (\texttt{gpt-5-mini} agent, \texttt{qwen3-235b} user simulator, \texttt{DeepSeek-V4-Flash} judge, $n{=}342$ per condition).
  \textbf{C:}~=~critical-severity count; \textbf{M:}~=~minor-severity count.
  Counts are judge-annotated error instances pooled over all simulations; a simulation may contribute several instances and an instance may carry several tags, so column sums can exceed $n$.
}
\label{tab:agent-tags-all-telecom}
\end{table*}

% ------ TELECOM -- AGENT ERROR TAGS -- PART B ------
\begin{table*}[p]
\centering
\footnotesize
\renewcommand{\arraystretch}{1.05}
\setlength{\tabcolsep}{4pt}
\begin{tabular}{@{}lccccc@{}}
\toprule
& \multicolumn{5}{c}{\textbf{L2 Tool --- Single-language}} \\
\cmidrule(lr){2-6}
\textbf{Error Tag} & VI & TH & ID & ZH & FIL \\
\midrule
  Guideline Violation & C:73 M:90 & C:69 M:77 & C:57 M:94 & C:77 M:61 & C:49 M:88 \\
  Hallucination & C:4 M:18 & C:4 M:11 & C:6 M:8 & C:2 M:14 & C:2 M:11 \\
  Inconsistent Behavior & C:6 M:29 & C:6 M:25 & C:3 M:29 & C:3 M:36 & C:7 M:36 \\
  Incorrect Interpretation & C:41 M:80 & C:17 M:91 & C:30 M:80 & C:37 M:74 & C:26 M:66 \\
  Interruption Error & C:0 M:0 & C:0 M:0 & C:0 M:0 & C:0 M:0 & C:0 M:0 \\
  Irrelevant Tool Call & C:1 M:11 & C:2 M:8 & C:2 M:27 & C:5 M:6 & C:2 M:11 \\
  Missed Required Action & C:140 M:154 & C:93 M:124 & C:104 M:133 & C:134 M:142 & C:96 M:147 \\
  Other & C:0 M:5 & C:1 M:14 & C:0 M:17 & C:2 M:5 & C:5 M:2 \\
  Premature Termination & C:25 M:1 & C:28 M:1 & C:28 M:4 & C:28 M:3 & C:23 M:1 \\
  Revealed Info Early & C:1 M:1 & C:1 M:3 & C:1 M:3 & C:1 M:1 & C:1 M:0 \\
  Tool Call Argument Error & C:0 M:2 & C:1 M:3 & C:1 M:4 & C:1 M:1 & C:1 M:7 \\
  Tool Call Schema Error & C:0 M:0 & C:0 M:0 & C:0 M:0 & C:0 M:1 & C:2 M:0 \\
  Wrong Sequence & C:12 M:15 & C:7 M:13 & C:10 M:14 & C:10 M:8 & C:9 M:9 \\
\bottomrule
\end{tabular}
\caption{%
  Agent error tag counts --- \textbf{Telecom} domain, L2 Tool (single-language) settings.
}
\label{tab:agent-tags-all-telecom-tool-l2}
\end{table*}

% ------ TELECOM -- AGENT ERROR TAGS -- PART C ------
\begin{table*}[p]
\centering
\footnotesize
\renewcommand{\arraystretch}{1.05}
\setlength{\tabcolsep}{4pt}
\begin{tabular}{@{}lcccc@{}}
\toprule
& \multicolumn{4}{c}{\textbf{L2 Tool --- Mixed-language}} \\
\cmidrule(lr){2-5}
\textbf{Error Tag} & Mix-2 & Mix-3 & Mix-4 & Mix-5 \\
\midrule
  Guideline Violation & C:80 M:83 & C:78 M:81 & C:70 M:75 & C:72 M:90 \\
  Hallucination & C:4 M:15 & C:3 M:14 & C:4 M:17 & C:3 M:14 \\
  Inconsistent Behavior & C:8 M:27 & C:9 M:29 & C:3 M:19 & C:7 M:26 \\
  Incorrect Interpretation & C:28 M:71 & C:36 M:98 & C:26 M:96 & C:40 M:80 \\
  Interruption Error & C:0 M:0 & C:0 M:0 & C:0 M:0 & C:0 M:0 \\
  Irrelevant Tool Call & C:2 M:14 & C:1 M:36 & C:0 M:6 & C:1 M:20 \\
  Missed Required Action & C:107 M:153 & C:100 M:189 & C:106 M:135 & C:118 M:118 \\
  Other & C:0 M:11 & C:1 M:27 & C:0 M:2 & C:1 M:6 \\
  Premature Termination & C:36 M:4 & C:28 M:4 & C:33 M:0 & C:37 M:2 \\
  Revealed Info Early & C:2 M:1 & C:0 M:0 & C:0 M:1 & C:1 M:4 \\
  Tool Call Argument Error & C:3 M:1 & C:2 M:3 & C:0 M:0 & C:0 M:6 \\
  Tool Call Schema Error & C:1 M:3 & C:0 M:2 & C:0 M:0 & C:0 M:0 \\
  Wrong Sequence & C:8 M:13 & C:7 M:16 & C:7 M:9 & C:12 M:18 \\
\bottomrule
\end{tabular}
\caption{%
  Agent error tag counts --- \textbf{Telecom} domain, L2 Tool (mixed-language) settings.
}
\label{tab:agent-tags-all-telecom-tool-mix}
\end{table*}

% ------ TELECOM -- AGENT ERROR TAGS -- PART D ------
\begin{table*}[p]
\centering
\footnotesize
\renewcommand{\arraystretch}{1.05}
\setlength{\tabcolsep}{4pt}
\begin{tabular}{@{}lccccc@{}}
\toprule
& \multicolumn{5}{c}{\textbf{L2 Domain}} \\
\cmidrule(lr){2-6}
\textbf{Error Tag} & VI & TH & ID & ZH & FIL \\
\midrule
  Guideline Violation & C:117 M:104 & C:106 M:97 & C:98 M:91 & C:78 M:81 & C:106 M:115 \\
  Hallucination & C:7 M:16 & C:5 M:11 & C:5 M:19 & C:5 M:9 & C:3 M:11 \\
  Inconsistent Behavior & C:10 M:37 & C:9 M:47 & C:8 M:44 & C:8 M:31 & C:12 M:38 \\
  Incorrect Interpretation & C:50 M:69 & C:54 M:95 & C:53 M:77 & C:29 M:71 & C:54 M:85 \\
  Interruption Error & C:0 M:0 & C:0 M:0 & C:0 M:0 & C:0 M:0 & C:0 M:0 \\
  Irrelevant Tool Call & C:3 M:19 & C:2 M:11 & C:1 M:8 & C:0 M:4 & C:3 M:15 \\
  Missed Required Action & C:151 M:96 & C:166 M:136 & C:178 M:108 & C:111 M:119 & C:163 M:111 \\
  Other & C:0 M:15 & C:1 M:9 & C:1 M:7 & C:0 M:7 & C:1 M:2 \\
  Premature Termination & C:30 M:3 & C:39 M:2 & C:33 M:3 & C:27 M:3 & C:33 M:5 \\
  Revealed Info Early & C:3 M:4 & C:1 M:4 & C:0 M:5 & C:0 M:0 & C:2 M:2 \\
  Tool Call Argument Error & C:2 M:7 & C:2 M:7 & C:2 M:3 & C:2 M:5 & C:3 M:11 \\
  Tool Call Schema Error & C:2 M:1 & C:0 M:0 & C:1 M:3 & C:0 M:0 & C:5 M:4 \\
  Wrong Sequence & C:17 M:13 & C:16 M:12 & C:23 M:21 & C:12 M:19 & C:13 M:27 \\
\bottomrule
\end{tabular}
\caption{%
  Agent error tag counts --- \textbf{Telecom} domain, L2 Domain settings.
}
\label{tab:agent-tags-all-telecom-domain}
\end{table*}

% ------ TELECOM -- USER ERROR TAGS -- PART A ------
\begin{table*}[p]
\centering
\footnotesize
\renewcommand{\arraystretch}{1.05}
\setlength{\tabcolsep}{4pt}
\begin{tabular}{@{}lcccccc@{}}
\toprule
& \textbf{Eng.~Baseline} & \multicolumn{5}{c}{\textbf{L2 Interaction}} \\
\cmidrule(lr){2-2}\cmidrule(lr){3-7}
\textbf{Error Tag} & EN & VI & TH & ID & ZH & FIL \\
\midrule
  Guideline Violation & C:25 M:42 & C:10 M:93 & C:30 M:77 & C:30 M:53 & C:30 M:82 & C:54 M:105 \\
  Hallucination & C:52 M:85 & C:34 M:96 & C:47 M:67 & C:68 M:74 & C:62 M:85 & C:29 M:68 \\
  Inconsistent Behavior & C:24 M:194 & C:19 M:207 & C:26 M:186 & C:33 M:139 & C:40 M:197 & C:52 M:194 \\
  Incorrect Interpretation & C:1 M:50 & C:6 M:30 & C:2 M:36 & C:8 M:49 & C:7 M:38 & C:3 M:31 \\
  Interruption Error & C:0 M:0 & C:0 M:0 & C:0 M:0 & C:0 M:0 & C:0 M:0 & C:0 M:0 \\
  Irrelevant Tool Call & C:0 M:0 & C:0 M:1 & C:0 M:1 & C:0 M:4 & C:0 M:0 & C:0 M:0 \\
  Missed Required Action & C:3 M:24 & C:3 M:9 & C:6 M:16 & C:1 M:22 & C:3 M:12 & C:2 M:29 \\
  Other & C:1 M:10 & C:1 M:9 & C:0 M:13 & C:1 M:9 & C:0 M:28 & C:0 M:15 \\
  Premature Termination & C:13 M:16 & C:9 M:24 & C:12 M:20 & C:10 M:18 & C:14 M:21 & C:17 M:16 \\
  Revealed Info Early & C:0 M:0 & C:0 M:0 & C:0 M:2 & C:0 M:1 & C:1 M:3 & C:0 M:3 \\
  Tool Call Argument Error & C:0 M:3 & C:2 M:7 & C:0 M:2 & C:2 M:1 & C:1 M:0 & C:1 M:2 \\
  Tool Call Schema Error & C:0 M:1 & C:5 M:6 & C:0 M:0 & C:0 M:2 & C:0 M:8 & C:0 M:8 \\
  Wrong Sequence & C:0 M:0 & C:0 M:0 & C:0 M:11 & C:4 M:6 & C:2 M:5 & C:0 M:4 \\
\bottomrule
\end{tabular}
\caption{%
  User error tag counts --- \textbf{Telecom} domain, English baseline and L2 Interaction settings
  (\texttt{gpt-5-mini} agent, \texttt{qwen3-235b} user simulator, \texttt{DeepSeek-V4-Flash} judge, $n{=}342$ per condition).
  \textbf{C:}~=~critical-severity count; \textbf{M:}~=~minor-severity count.
  For the simulated user, \textbf{C:} pools \texttt{critical\_helped} and \texttt{critical\_hindered}.
  Counts are judge-annotated error instances pooled over all simulations; a simulation may contribute several instances and an instance may carry several tags, so column sums can exceed $n$.
}
\label{tab:user-tags-all-telecom}
\end{table*}

% ------ TELECOM -- USER ERROR TAGS -- PART B ------
\begin{table*}[p]
\centering
\footnotesize
\renewcommand{\arraystretch}{1.05}
\setlength{\tabcolsep}{4pt}
\begin{tabular}{@{}lccccc@{}}
\toprule
& \multicolumn{5}{c}{\textbf{L2 Tool --- Single-language}} \\
\cmidrule(lr){2-6}
\textbf{Error Tag} & VI & TH & ID & ZH & FIL \\
\midrule
  Guideline Violation & C:33 M:52 & C:36 M:61 & C:16 M:74 & C:19 M:64 & C:35 M:58 \\
  Hallucination & C:45 M:91 & C:48 M:88 & C:51 M:78 & C:49 M:55 & C:32 M:60 \\
  Inconsistent Behavior & C:17 M:186 & C:47 M:175 & C:15 M:180 & C:22 M:160 & C:23 M:135 \\
  Incorrect Interpretation & C:4 M:30 & C:6 M:36 & C:5 M:49 & C:6 M:52 & C:4 M:46 \\
  Interruption Error & C:0 M:0 & C:0 M:0 & C:0 M:0 & C:0 M:0 & C:0 M:0 \\
  Irrelevant Tool Call & C:0 M:1 & C:0 M:1 & C:0 M:8 & C:0 M:3 & C:0 M:10 \\
  Missed Required Action & C:2 M:11 & C:2 M:30 & C:2 M:21 & C:5 M:16 & C:9 M:11 \\
  Other & C:0 M:15 & C:1 M:3 & C:0 M:12 & C:0 M:32 & C:0 M:11 \\
  Premature Termination & C:5 M:23 & C:3 M:14 & C:12 M:20 & C:8 M:15 & C:11 M:14 \\
  Revealed Info Early & C:0 M:1 & C:0 M:1 & C:0 M:0 & C:0 M:4 & C:0 M:15 \\
  Tool Call Argument Error & C:0 M:10 & C:0 M:5 & C:1 M:9 & C:0 M:1 & C:0 M:2 \\
  Tool Call Schema Error & C:0 M:3 & C:0 M:6 & C:0 M:5 & C:0 M:2 & C:0 M:4 \\
  Wrong Sequence & C:0 M:7 & C:1 M:2 & C:0 M:6 & C:2 M:3 & C:0 M:3 \\
\bottomrule
\end{tabular}
\caption{%
  User error tag counts --- \textbf{Telecom} domain, L2 Tool (single-language) settings.
}
\label{tab:user-tags-all-telecom-tool-l2}
\end{table*}

% ------ TELECOM -- USER ERROR TAGS -- PART C ------
\begin{table*}[p]
\centering
\footnotesize
\renewcommand{\arraystretch}{1.05}
\setlength{\tabcolsep}{4pt}
\begin{tabular}{@{}lcccc@{}}
\toprule
& \multicolumn{4}{c}{\textbf{L2 Tool --- Mixed-language}} \\
\cmidrule(lr){2-5}
\textbf{Error Tag} & Mix-2 & Mix-3 & Mix-4 & Mix-5 \\
\midrule
  Guideline Violation & C:15 M:49 & C:13 M:52 & C:15 M:77 & C:12 M:56 \\
  Hallucination & C:48 M:93 & C:40 M:95 & C:38 M:69 & C:74 M:83 \\
  Inconsistent Behavior & C:28 M:158 & C:32 M:209 & C:23 M:169 & C:14 M:184 \\
  Incorrect Interpretation & C:4 M:38 & C:7 M:35 & C:4 M:39 & C:4 M:29 \\
  Interruption Error & C:0 M:0 & C:0 M:0 & C:0 M:0 & C:0 M:0 \\
  Irrelevant Tool Call & C:0 M:0 & C:0 M:1 & C:0 M:3 & C:0 M:0 \\
  Missed Required Action & C:3 M:22 & C:2 M:8 & C:4 M:17 & C:1 M:21 \\
  Other & C:1 M:9 & C:0 M:16 & C:1 M:10 & C:1 M:10 \\
  Premature Termination & C:11 M:15 & C:12 M:17 & C:15 M:14 & C:5 M:16 \\
  Revealed Info Early & C:1 M:1 & C:1 M:1 & C:0 M:0 & C:0 M:0 \\
  Tool Call Argument Error & C:1 M:4 & C:4 M:8 & C:0 M:1 & C:0 M:1 \\
  Tool Call Schema Error & C:0 M:3 & C:0 M:1 & C:0 M:5 & C:0 M:1 \\
  Wrong Sequence & C:0 M:10 & C:0 M:1 & C:0 M:5 & C:0 M:2 \\
\bottomrule
\end{tabular}
\caption{%
  User error tag counts --- \textbf{Telecom} domain, L2 Tool (mixed-language) settings.
}
\label{tab:user-tags-all-telecom-tool-mix}
\end{table*}

% ------ TELECOM -- USER ERROR TAGS -- PART D ------
\begin{table*}[p]
\centering
\footnotesize
\renewcommand{\arraystretch}{1.05}
\setlength{\tabcolsep}{4pt}
\begin{tabular}{@{}lccccc@{}}
\toprule
& \multicolumn{5}{c}{\textbf{L2 Domain}} \\
\cmidrule(lr){2-6}
\textbf{Error Tag} & VI & TH & ID & ZH & FIL \\
\midrule
  Guideline Violation & C:13 M:42 & C:17 M:46 & C:11 M:59 & C:23 M:85 & C:22 M:85 \\
  Hallucination & C:39 M:48 & C:49 M:56 & C:57 M:54 & C:37 M:61 & C:52 M:47 \\
  Inconsistent Behavior & C:18 M:146 & C:16 M:151 & C:20 M:140 & C:10 M:115 & C:20 M:142 \\
  Incorrect Interpretation & C:3 M:30 & C:7 M:34 & C:8 M:25 & C:6 M:28 & C:7 M:42 \\
  Interruption Error & C:0 M:0 & C:0 M:0 & C:0 M:0 & C:0 M:0 & C:0 M:0 \\
  Irrelevant Tool Call & C:0 M:1 & C:0 M:0 & C:0 M:1 & C:0 M:3 & C:0 M:3 \\
  Missed Required Action & C:6 M:8 & C:2 M:12 & C:1 M:9 & C:3 M:14 & C:2 M:9 \\
  Other & C:1 M:11 & C:0 M:33 & C:0 M:20 & C:1 M:8 & C:0 M:27 \\
  Premature Termination & C:11 M:23 & C:17 M:12 & C:14 M:29 & C:9 M:18 & C:11 M:31 \\
  Revealed Info Early & C:1 M:0 & C:1 M:1 & C:1 M:2 & C:0 M:2 & C:0 M:1 \\
  Tool Call Argument Error & C:0 M:5 & C:0 M:2 & C:0 M:1 & C:0 M:3 & C:0 M:2 \\
  Tool Call Schema Error & C:0 M:0 & C:0 M:1 & C:0 M:22 & C:1 M:1 & C:0 M:2 \\
  Wrong Sequence & C:0 M:2 & C:1 M:1 & C:0 M:1 & C:0 M:1 & C:0 M:1 \\
\bottomrule
\end{tabular}
\caption{%
  User error tag counts --- \textbf{Telecom} domain, L2 Domain settings.
}
\label{tab:user-tags-all-telecom-domain}
\end{table*}

\end{document}